\documentclass[10pt,letterpaper]{article}
\usepackage[top=0.85in,left=2.75in,footskip=0.75in]{geometry}

\usepackage{amsmath,amssymb}

\usepackage{changepage}

\usepackage{textcomp,marvosym}

\usepackage{cite}

\usepackage{nameref,hyperref}

\usepackage[nopatch=eqnum]{microtype}
\DisableLigatures[f]{encoding = *, family = * }

\usepackage[table]{xcolor}

\usepackage{array}

\usepackage[most]{tcolorbox}
\usepackage{subfigure}

\newcolumntype{+}{!{\vrule width 2pt}}

\newlength\savedwidth

\raggedright
\usepackage[aboveskip=1pt,labelfont=bf,labelsep=period,justification=raggedright,singlelinecheck=off]{caption}

\makeatletter
\renewcommand{\@biblabel}[1]{\quad#1.}
\makeatother

\usepackage{lastpage,fancyhdr,graphicx}
\usepackage{epstopdf}
\fancyheadoffset[L]{2.25in}
\fancyfootoffset[L]{2.25in}
\begin{document}
\vspace*{0.2in}

\begin{flushleft}
{\Large
\textbf\newline{Nonlinear Dimensionality Reduction with Diffusion Maps in Practice} % Please use "sentence case" for title and headings (capitalize only the first word in a title (or heading), the first word in a subtitle (or subheading), and any proper nouns).
}
\newline
\\
S\"onke Beier\textsuperscript{1\Yinyang},
Paula Pirker-D\'{i}az\textsuperscript{1\Yinyang},
Friedrich Pagenkopf\textsuperscript{2\Yinyang},
Karoline Wiesner\textsuperscript{1*}

\bigskip
\textbf{1} Institute of Physics and Astronomy, University of Potsdam, D-14476 Potsdam, Germany
\\
\textbf{2} University of Leipzig, D-04103 Leipzig, Germany
\\

\bigskip

% Insert additional author notes using the symbols described below. Insert symbol callouts after author names as necessary.
% 
% Remove or comment out the author notes below if they aren't used.
%
% Primary Equal Contribution Note
\Yinyang These authors contributed equally to this work.

% Additional Equal Contribution Note
% Also use this double-dagger symbol for special authorship notes, such as senior authorship.

% Current address notes% change symbol to "\textcurrency a" if more than one current address note
% \textcurrency b Insert second current address 
% \textcurrency c Insert third current address


% Group/Consortium Author Note

% Use the asterisk to denote corresponding authorship and provide email address in note below.
* Corresponding author: karoline.wiesner@uni-potsdam.de

\end{flushleft}
\section*{Abstract}
Diffusion Map is a spectral dimensionality reduction technique which is able to uncover nonlinear submanifolds in high-dimensional data.
And, it is increasingly applied across a wide range of scientific disciplines, such as biology, engineering, and social sciences.
But data preprocessing, parameter settings and component selection have a significant influence on the resulting manifold, something which has not been comprehensively discussed in the literature so far. We provide a practice oriented review of the Diffusion Map technique, illustrate pitfalls and showcase a recently introduced technique for identifying the most relevant components. 
Our results show that the first components are not necessarily the most relevant ones.

% Please keep the Author Summary between 150 and 200 words
% Use first person. PLOS ONE authors please skip this step. 
% Author Summary not valid for PLOS ONE submissions.   
%\section*{Author summary}
%Lorem ipsum dolor sit amet, consectetur adipiscing elit. Curabitur eget porta erat. Morbi consectetur est vel gravida pretium. Suspendisse ut dui eu ante cursus gravida non sed sem. Nullam sapien tellus, commodo id velit id, eleifend volutpat quam. Phasellus mauris velit, dapibus finibus elementum vel, pulvinar non tellus. Nunc pellentesque pretium diam, quis maximus dolor faucibus id. Nunc convallis sodales ante, ut ullamcorper est egestas vitae. Nam sit amet enim ultrices, ultrices elit pulvinar, volutpat risus.

%\linenumbers
\tableofcontents
% Use "Eq" instead of "Equation" for equation citations.
\section{Introduction}
%situation
In an increasingly connected and digitized  
world, the volume of complex, high-dimensional
data is growing rapidly across a broad range of disciplines, from biology, neuroscience, chemistry and physics, to social sciences. 
%
%problem
High dimensionality offers a deeper insight and potential of description, but it challenges the analysis, the visualization and the interpretation of the data.
%
%solution
Dimensionality reduction methods serve as essential tools for addressing these challenges, making such high-dimensional data more manageable by projecting it onto an interpretable low-dimensional space while preserving its essential structure.

Standard approaches such as Principal Component Analysis (PCA) \cite{Pearson1901,Hotelling1933} and the Multidimensional Scaling method \cite{Torgerson1952,Ek2021} rely on linear transformations. They are very efficient, but are not able to capture nonlinear relationships prevalent in many real-world data sets. 
To capture such nonlinear patterns, non-linear dimensionality reduction methods need to be used. Here, spectral methods are particularly widespread. Examples are the Locally Linear Embedding \cite{roweis2000}, the Isomap \cite{Tenenbaum2000}, Spectral Clustering or Embedding \cite{Ng2001}, Kernel PCA \cite{schoelkopf1997}, Laplacian Eigenmaps \cite{Belkin2001} and the Diffusion Map \cite{lafon2004,Nadler2005,Coifman2006}. Notably, probabilistic perspectives on this type of technique had already been proposed earlier \cite{Meila2000,Meila2001}. However, earlier developments of this class of methods can also be found in the literature \cite{Scott1990,Weiss1999}. The core concept behind all these methods is rooted in the manifold hypothesis, which suggests that real-world datasets typically lie on a non-linear low-dimensional manifold embedded within the original high-dimensional space. Consequently, dimensionality reduction methods are also called manifold learning techniques. They aim to uncover this underlying structure, thereby enabling the construction of meaningful low-dimensional representations. Due to the conceptual similarities among these methods, several works have provided unified frameworks for nonlinear spectral techniques \cite{ham2004,Bengio2004,Strange2014,Ghojogh2023}.

The Diffusion Map method is a spectral dimensionality reduction method that offers a probabilistic interpretation based on the physics of diffusion \cite{Nadler2005}. 
And, it is increasingly applied across a wide range of scientific disciplines, including biology \cite{Xu2010,Fahimipour2020}, chemistry \cite{Ferguson2010, Kim2015,trstanova_2020}, engineering \cite{Bohn2013,Schwartz2014,Chen2016,Ghafourian2020}, geophysics \cite{Rabin2016,bregman_2021} and social sciences \cite{Le2013,Barter2019,Levin2021, Xiu2023,Xiu2024,Zeng2024,paula2024,Beier2025}.
%our problem 
But 
%However, a throughout and practice oriented review of the Diffusion Map technique is absent from the literature. In particular, we found the
data preprocessing, parameter settings and component selection have a significant influence on the resulting manifold, something which is seldomdiscussed in detail in the existing literature. Therefore, we provide a practice oriented review of the Diffusion Map technique, illustrating pitfalls and highlighting a recently introduced technique for selecting the most relevant components.

%here we come

%The paper pursues two main goals: (1) reviewing the Diffusion Map technique with detailed instructions for its use in practice and (2) introducing a quantitative method to assess the relevance of Diffusion Map components, based on neural networks.

The article is organized as follows. We first review the Diffusion Map method and  the associated computational algorithm. Next, we study its practical  properties -- the effect of different parameter settings and  the influence of data pre-processing -- with help of a standard example data set. We then discuss reasons when and why the most relevant component is not necessarily the first one, and end with reviewing a technique for quantifying the relevance of each component. %The conclusions are provided together with some perspectives in the final section.

%Specifically, we examine the effects of scaling individual variables, normalization procedures, and the presence of redundant variables, as well as the selection of the different Diffusion Map parameters. Furthermore, we explore methods for quantitatively assessing the relevance of Diffusion Map components. For that purpose, we propose an approach based on neural networks.\\
%nochmal schreiben wo was ist? Aber erst am Ende wenn alles fertig ist - Ja! :)

\section{Diffusion Map: the method}

In the following, we will review the core ideas behind nonlinear spectral dimensionality reduction techniques in general and Diffusion Map specifically. \\

\begin{tcolorbox}[colback=gray!10, colframe=black, title=Summary of the Diffusion Map algorithm]
{\small \textsc{INPUT:}} High-dimensional dataset with $n$ data points: $\left\{x_i \right\}_{i=1}^n, x_i \in \mathbb{R}^p$, where $p\ll n$\\

1. Definition of a symmetric and positive semi-definite kernel function, $k(x_i, x_j)$, and creation of the kernel matrix, $K_{ij}=k(x_i, x_j)$, see Eq. \ref{kernel} for the Gaussian kernel definition. \\
2. Anisotropic kernel construction, $L=D^{-\alpha}KD^{-\alpha}$, where $D_{ii}=\sum_j K_{ij}$.\\
3. Computation of the transition matrix (normalized graph Laplacian) by normalizing the kernel, $M = \tilde{D}^{-1}L$, where $\tilde{D}_{ii}=\sum_j L_{ij}$. \\
4. Spectral decomposition of matrix $M$ or of its symmetric version, $M_s=\tilde{D}^{\frac{1}{2}}M\tilde{D}^{-\frac{1}{2}}$. \\
5. Definition of the family of Diffusion Maps $\{\Psi^{(t)}(x)\}_{t\in\mathbb{N}}$, taking the eigenvalues ($\{\lambda_i\}$) and the right eigenvectors ($\{\psi_i\}$) obtained in step 3,
    \begin{equation}
    \label{eq:diffusion_map}
        \Psi^{(t)}(x) =\begin{pmatrix}
                    \Psi_1(x;t)\\
                    \Psi_2(x;t)\\
                    \vdots \\
                    \Psi_{k}(x;t)
                    \end{pmatrix} = \begin{pmatrix}
                    \lambda_1^t \psi_1(x)\\
                    \lambda_2^t \psi_2(x)\\
                    \vdots \\
                    \lambda_{k}^t \psi_{k}(x)
                    \end{pmatrix}
    \end{equation}
Here, $k$ is the number of diffusion components we decide to keep.
For $k=n-1$ we obtain the full set of diffusion components. \\
6. Selection of optimal combination of diffusion components to include in the embedding $\{\Psi_i(x;t)\}$. \\

{\small \textsc{OUTPUT:}} Diffusion Map embedding $\{\Psi_i(x;t)\}$.
\end{tcolorbox}

\subsection{Kernel definition. Creating a neighborhood graph}

All spectral dimensionality reduction methods start by finding the neighborhood information of the data \cite{Meila2024}. This process is best understood as constructing a neighborhood graph, where the nodes represent the individual data points and the edge weights represent their similarities. As a result, the graph encodes the local topological properties of the data \cite{Meila2024}. \\
As a similarity measure, the standard choice is the Euclidean distance $||x_i-x_j||$: distant data points are less similar than close by data points. As we only provide the distance of the data points in the algorithm and not their locations, the resulting neighborhood graph is independent of rotations, translations, or reflections of the original data \cite{Nadler2007}.\\

Now we add locality to the similarity measure: %in order to ensure the accurate recognition of manifolds, it is essential to adjust the parameters such that only data points within a local neighborhood are connected. This adjustment is necessary to ensure that the method can effectively follow the respective manifold.
%We therefore apply a \emph{kernel function} $k$ to build a kernel matrix
%\begin{equation}
%    K_{ij} = k(x_i, x_j)
%\end{equation}
Applying a \emph{kernel function} $k$
A widely used choice of kernel function is the Gaussian neighborhood kernel,% which is determined by the width parameter $\epsilon$. The Gaussian kernel matrix $K_{ij}$, which stores the local similarities by encoding the graph, is given by:
    \begin{equation}\label{kernel}
    K_{ij} = k(x_i, x_j) =  exp \left( -\frac{\lVert x_i-x_j \rVert^2}{\epsilon}\right)
    \end{equation}
Note that different conventions for the width parameter exist in the literature, where the denominator is e.g. $\epsilon^2$, $2 \epsilon^2$ or $4\epsilon$ \cite{Nadler2007, Ghojogh2023, shan2022semigroup}.
Other kernels are also possible; for instance, \cite{Barter2019} uses a reciprocal function. Other methods can also be used to ensure locality \cite{Luxburg2007,Meila2024}.

By only considering the $N$ nearest data points \cite{Izenman2012,Barter2019,Fahimipour2020,Xiu2023}, the resulting neighborhood graph is a symmetric and positive semi-definite matrix with entries

\begin{equation}
    K_{ij} = \begin{cases}
    k(x_i, x_j), & \text{if $x_i, x_j$ are}\\ & \text{$N$-nearest neighbors} \\
    0, & \text{else}
    \end{cases}
\end{equation}
This is a sparse matrix, which makes the subsequent computations very efficient \cite{DeSilva2002, Luxburg2007, VanDerMaaten2009a, Barter2019, Meila2024}, but the choice of $N$ can have a significant impact on the results \cite{Beier2025}.

\begin{figure}
    \centering
\includegraphics[width=1\linewidth]{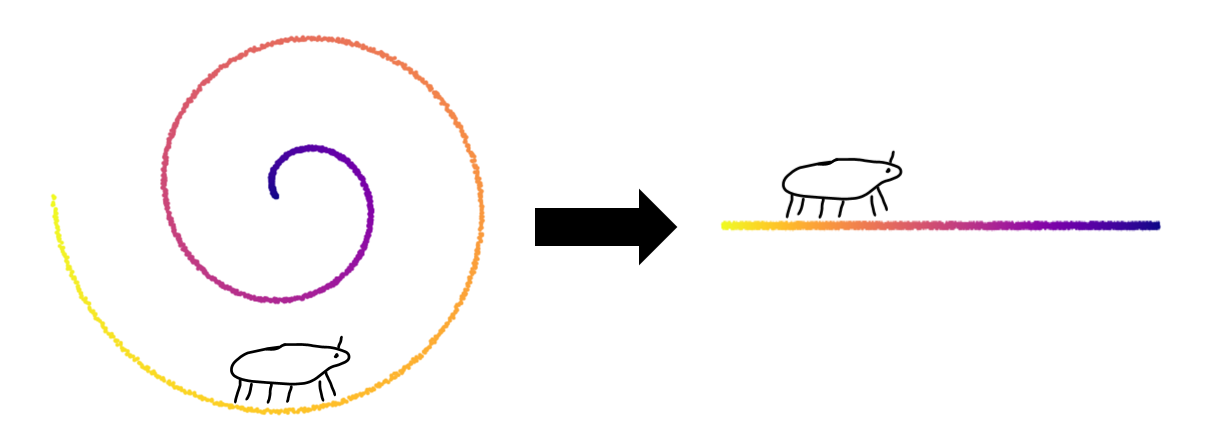}
    \caption{\textbf{Illustrating the idea of nonlinear dimensionality reduction:} On the left the data points lie on a one-dimensional manifold (spiral) in a 2 dimensional space. The idea of nonlinear dimensional reduction is to parameterize the underlying manifold. The procedure can be compared to a small insect located at one data point that only recognizes its immediate environment, but not the global appearance of the data in a higher-dimensional sense. If it moves along the manifold, it can put together the local information to create the low-dimensional representation (right). The color can be seen as the natural parameter of the manifold. The illustration is inspired by \cite{Ghojogh2023}.}
    \label{fig:illustration_local}
\end{figure}

%But why is the neighborhood kernel important? If two points are close to each other, the Euclidean distance gives us useful information; the global ‘long-range’ similarities given by the Euclidean distance are not important and trustworthy \cite{lafon2004,Luxburg2007,Porte2008} and should not be preserved. The locality is crucial to be able to recognize the non-linear global structures. This is also the difference to linear spectral methods such as PCA which do not use local, but global distances and similarity matrices.\\
%In the context of the probabilistic interpretation of the Diffusion Map, the defined similarity by the symmetric and positive semi-definite kernel described by $K_{ij}$ is related to the probability of transition in one time step from data point $x_i$ to $x_j$. In this description, the kernel is constructed so that transitions between nearest neighbors are the most probable. By doing so, we ensure that the transition matrix encapsulates the local geometric information of the data set. And this is essential, as locality preservation is crucial for recognizing the underlying non-linear global structures.
Consequently, the global information can be reconstructed by focusing on the local information as illustrated in figure \ref{fig:illustration_local}.

\subsection{Anisotropic kernel construction}

After defining the kernel matrix $K$, an anisotropic version of it can be computed. We introduce $L = D^{-\alpha}KD^{-\alpha}$, where $D$ is the diagonal matrix with $D_{ii} = \sum_{j=1}^n K_{ij}$.
$\alpha \geq 0$ is a parameter that controls how strongly density fluctuations of data points affect the outcome of the Diffusion Map \cite{Coifman2006}. 
One can set $\alpha = 0$, which equates to setting $L = K$, thus skipping this normalization step. In that case, the influence of the density is maximal. This corresponds to the classical normalized graph Laplacian \cite{Coifman2006}.
However, as argued in \cite{Coifman2006}, there are good reasons to set $\alpha=1$ or $\alpha=\frac{1}{2}$.
In that case, since $D_{ii}$ is an estimate of the local density at point $x_i$, this normalization amounts to weakening the couplings in dense regions and strengthening them in sparse regions, thus making the embedding more resilient to varying sample densities.
We will come back to the concrete meaning of $\alpha$ when discussing the asymptotics of the Diffusion Map (section \ref{asymptoticssection}) and its effects in section \ref{alphasection}.

\subsection{Transition matrix computation}

To give the neighborhood graph a probabilistic interpretation, a different normalization is applied to construct a random walk out of $L$.
By normalizing the matrix $L$ with the diagonal matrix described by $\tilde{D}_{ii} = \sum_{j=1}^n L_{ij}$, one obtains a row-stochastic Markov matrix $M = \tilde{D}^{-1}L$ representing transition probabilities between data points. It is important to mention that, for a large enough $\epsilon$, $M$ is the transition matrix of an irreducible Markov chain.
We note that $M$ is related to its symmetric matrix $M_s = \tilde{D}^{1/2}M\tilde{D}^{-1/2} = \tilde{D}^{-1/2}L\tilde{D}^{-1/2}$ by a change of basis.
It therefore has the same spectral decomposition and that makes $M_s$ very suitable for easing the computation of the spectral decomposition, which is the next step.

We are able to let the Markov process evolve forward in time. The transition probabilities after $t$ discrete time steps are given by the $t$-th power of the Markov matrix $M^t$, where the entry $(M^t)_{ij}$ represents the probability of transitioning from data point $i$ to data point $j$ in $t$ time steps \cite{Coifman2006}. %However, we will see in the results section that the parameter $t$ does not influence the qualitative outcome of the Diffusion Map embedding.\\

\subsection{Spectral decomposition}

In this step, the local information stored in the matrices $M$ or $M_s$, respectively, is reassembled into a global representation of the manifold through the spectral decomposition \cite{lafon2004,ham2004,Saul2003,Wu2018}. This procedure is often described as the preservation of the global structure through local fitting. \cite{Ghojogh2023}.\\

%As mentioned in \cite{Nadler2005} we do a spectral decomposition of the symmetric matrix $M_s$, which should improve the performance of the algorithm in comparison to using $M$ directly\cite{Ng2001}. 
Since the time parameter $t$ only influences the result of the eigendecomposition  by raising the eigenvalues to the power $t$, while the eigenvectors remain unaffected, we can diagonalize $M^t$ for any $t$ by diagonalizing $M_s$. We get the normalized eigenvectors of $M_s$, denoted as ${v_i}$. $M$ and $M_s$ have the same eigenvalues ${\lambda_i}$, which we can sort by magnitude. Since $M$ is the transition matrix of an irreducible Markov chain, we have $1=\lambda_0 > \lvert \lambda_1 \rvert \geq \lvert \lambda_2 \rvert \geq\dots$ \cite{Coifman2006}. The left eigenvectors ${\phi_i}$ and the right eigenvectors ${\psi_i}$ of $M$ can be calculated by $\phi_i = \frac{1}{\sqrt{tr(D)}}v_i D^{1/2}$ and $\psi_i = \sqrt{tr(D)}v_i D^{-1/2}$.\\

The eigenvector corresponding to eigenvalue $\lambda_0=1$ is worth mentioning. 
Given that $M$ is a stochastic matrix, it satisfies $M\overrightarrow{1}= \overrightarrow{1}$, where $\overrightarrow{1}$ is a vector of ones. Therefore, the right eigenvector corresponding to eigenvalue 1 is $\psi_0 = \overrightarrow{1}$. $\psi_0$ is not included in the Diffusion Map construction, as it does not distinguish different nodes of the graph.
The left eigenvector corresponding to eigenvalue 1, $\phi_0$, is the stationary distribution of the Markov chain defined by $M$ \cite{Nordstrom2005FINITEMC}. \\

\subsection{Diffusion Map definition}

With the decomposition of the transition matrix $M$ we now have $n$ eigenvalues and eigenvectors, which embed the $n$ original data points. The family of Diffusion Maps is the set $\{\Psi^{(t)}(x)\}_{t\in\mathbb{N}}$, such that %, is defined by taking the eigenvalues ($\{\lambda_i\}$) and the right eigenvectors ($\{\psi_i\}$) obtained in the last step. 
for each choice of $t$ we obtain a Diffusion Map defined as 
\begin{equation}\label{eq-dm-def}
    \Psi^{(t)}(x) = \begin{pmatrix}
                \Psi_1(x;t)\\
                \Psi_2(x;t)\\
                \vdots \\
                \Psi_s(x;t)
                \end{pmatrix} =
                \begin{pmatrix}
                \lambda_1^t \psi_1(x)\\
                \lambda_2^t \psi_2(x)\\
                \vdots \\
                \lambda_k^t \psi_{k}(x)
                \end{pmatrix},
\end{equation}
where $k \leq n-1$ is the number of components we decide to retain (see Summary Box).
%Each dimension of the embedding is called a diffusion component, and the whole map embeds the original data into a new lower-dimensional diffusion space of $k$ dimensions.
Typically, very small values of $k$ are chosen; of course, to achieve a dimensionality reduction, one should choose $k<p$.

\subsection{Diffusion distance}
The geometry of the manifold is characterized by the so-called \textit{diffusion distance}, which is defined by 

\begin{align}
        D_t^2(x_0, x_1)&=  \lVert p(t, y|x_0)-p(t, y|x_1) \rVert^2_w \\
        &= \sum_{y} (p(t, y|x_0)-p(t, y|x_1))^2 w(y),
\end{align}

where $p(t,y|x_i)$ is the probability distribution of a random walk landing at point $y$ at time $t$, given a starting location $x_i$ at time $t=0$. Equivalently, $p(t,y|x_i)=e_iM^t$, where $e_i$ denotes the row vector with all zeros except a single 1 in the $i$-th position. $w(y)$ is the weight function, which corresponds to the inverse of the stationary distribution, $1/\phi_0(y)$.
%\textcolor{red}{The \textit{diffusion distance} is the $L^2$ distance between the probability bumps at time $t$ centered around $x_0$ and $x_1$, respectively.}
%The interpretation is therefore that it measures how different the diffusion processes are in velocity, starting at different points.
The key property of the Diffusion Map is that, the \textit{diffusion distance}, $D_t^2$, in the original data space equals the Euclidean distance in the Diffusion Map space \cite{Coifman2006, Nadler2007}. When all $n-1$ diffusion components are included:

\begin{align}\label{eq-dm-difdis}
        D_t^2(x_0, x_1)&=\sum_{j=1}^{n-1} \lambda_j^{2t}(\psi_j(x_0)-\psi_j(x_1))^2 \\ &= \lVert \Psi^{(t)}(x_0)-\Psi^{(t)}(x_1) \rVert^2
\end{align}

The Euclidean distance between two points in the original space does not depend on the position of the rest of data points. In contrast, the \textit{diffusion distance} depends on all possible paths connecting the two points considered, giving a proximity measure depending on the connectivity of the data points in the neighborhood graph (see Eq. \ref{eq-dm-difdis}). As remarked in \cite{Coifman2006}, this highlights the notion of clustering: points are regarded as closer when they are strongly connected in the graph. 

By using fewer diffusion components, we get lower dimensions at the cost of a lower relative accuracy of the embedding in the sense of the \textit{diffusion distance}. This error is bounded by a term proportional to $\lambda_k^{2t}$ \cite{Nadler2007}, which is why the spectrum $\lambda^t$ is commonly used to evaluate the accuracy of Diffusion Maps \cite{Coifman2006, Coifman2005, COIFMAN20052,Haghverdi2015}. We show in Section \ref{sectioncomponents} that the decay of $\lambda_k^{2t}$ is at best not helpful and at worst misleading for this purpose (see Fig. \ref{fig:swissroll_spectrum}).
The discussion about how many diffusion components should be retained is quite brief in the literature, but it is often recommended to define a threshold $\delta>0$ and to keep the first $s(\delta,t)$ components, where
\begin{equation}
    s(\delta,t)=\max\{l\in \mathbb{N} \text{ such that } |\lambda_l|^t > \delta|\lambda_1|^t\},
\end{equation}
see references \cite{Coifman2006, Coifman2005, COIFMAN20052}. In addition, others argue that a large spectral gap after the eigenvalue $\lambda_l$ implies that considering the diffusion components up to $\psi_l$ typically yields a good approximation of diffusion distances \cite{Nadler2007,Haghverdi2015}. It is worth mentioning that notation varies depending on the source and $s$ is sometimes called $k$. 
We will come back to the component selection criteria in section \ref{sectioncomponents}. \\

Considering Eq. \ref{eq-dm-def} and the fact that $1=\lambda_0 > \lvert \lambda_1 \rvert \geq \lvert \lambda_2 \rvert \geq\dots\geq \lvert \lambda_n \rvert$, the larger the eigenvalue, the slower the convergence of the corresponding diffusion mode to the stationary state.
This means that the components corresponding to the largest eigenvalues (i.e. the first ones), correspond to the directions of slower diffusion and, consequently, are the ones retaining the most pronounced geometric features.

As noted in \cite{Nadler2007}, diffusion components can be redundant. For example, components can be %may not be relevant for analysis because they do not contribute additional information. Instead, they ``encode for the same geometrical or spatial direction of the manifold", representing merely higher-order 
polynomial functions of diffusion components. Removing these redundant diffusion components can be beneficial during data analysis \cite{Nadler2007}.\\
%Note that when computing the Diffusion Map multiple times, the resulting diffusion components may appear flipped. This is because eigenvectors of $M$ are only defined up to multiplication by a scalar, meaning that multiplying an eigenvector by $-1$ still yields a valid eigenvector.\\

\subsection{Asymptotics of the Diffusion Map}\label{asymptoticssection}

In \cite{Nadler2005, Nadler2007, Nadler2006}, the asymptotics of the Diffusion Map were analyzed.
In particular, the limiting cases $n\to\infty$ and $\epsilon\to 0$ are of interest to help interpret the results of the Diffusion Map.
In this section, we shall assume that data points are drawn i.i.d. from a continuous probability distribution $P$ which is supported on some compact manifold.
It helps to think of $P$ as being the Boltzmann distribution due to some potential $U$:
\begin{equation}
    P(x) = e^{-U(x)}.
\end{equation}
As the number of data points $n$ goes to infinity, the discrete Markov process approaches a process continuous in space and discrete in time, described by an operator $T$.
By also taking the limit $\epsilon \to 0$ we obtain a process which is also continuous in time and is described by the operator $\mathbf{H}$, the infinitesimal generator of $T$:
\begin{equation}
    \mathbf{H} = \lim_{\epsilon \to 0} \frac{T-I}{\epsilon}
\end{equation}
It was shown in \cite{Nadler2006} that
\begin{equation}\label{eq:asymptoticeq}
    \mathbf{H}\psi = \Delta \psi - 2(1-\alpha)\nabla \psi \cdot \nabla U.
\end{equation}

We see that the choice of $\alpha$ has a strong influence on the result when the density of data points is not constant on the manifold.
When $\alpha=0$, the influence of the density is maximal \cite{Coifman2006} and the infinitesimal generator $\mathbf{H}$ corresponds to a diffusion process in a potential $2U$\cite{Nadler2006}.
On the other hand, the choice $\alpha=1$ completely removes the influence of the density, since in that case, the infinitesimal generator is simply the Laplace-Beltrami operator, $\Delta \psi$ \cite{Coifman2006}.
This allows to recover only the geometry of the data manifold, without regard for the distribution of data points.
For $\alpha=\frac{1}{2}$ Eq. \ref{eq:asymptoticeq} turns into the Fokker-Planck equation of a diffusion process in potential $U$ \cite{Coifman2006}.

\subsubsection{Polynomial dependencies for one-dimensional manifolds}

It follows that the eigenvectors of the matrix $M$ (as defined in the step 3 of the Summary Box) can be interpreted as approximations of the eigenfunctions of $\mathbf{H}$ (Eq. \ref{eq:asymptoticeq}), if there are enough data points on the manifold \cite{Nadler2007}. 

\begin{figure*}[]
    \centering
    \includegraphics[width=\linewidth]{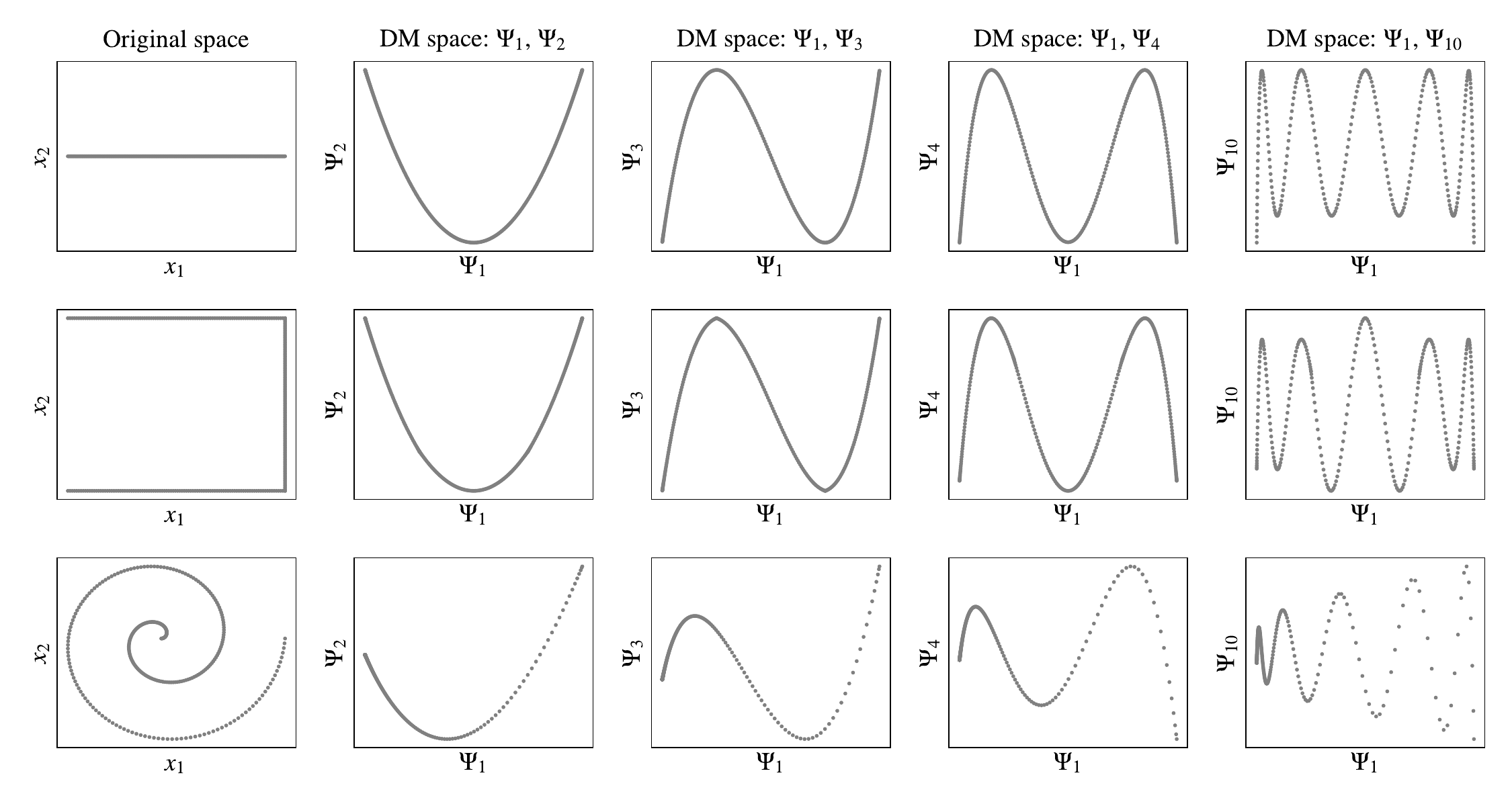}
    \caption{\textbf{Illustrating typical shape of Diffusion Map embedding for one dimensional non-intersecting datasets.} The first column displays the original datasets (consisting each of 300 data points) in the feature space. The subsequent columns show different diffusion components with respect to the first diffusion component $\Psi_1$ for each dataset (row).} 
\label{fig:polynomial_examples}
\end{figure*}

From \cite{Nadler2007} we know that Diffusion Map succeeds in finding a reasonable embedding of the data even in the simplest case, a one dimensional curve embedded in a high dimensional space:
\begin{quote}
    "\textit{\textbf{Theorem:} Consider data sampled }uniformly\textit{ from a non-intersecting smooth 1-D curve embedded in a high dimensional space. Then, in the limit of a large number of samples and small kernel width the first Diffusion Map coordinate gives a one-to-one parametrization of the curve.}"
\end{quote}
This theorem is illustrated in Figure \ref{fig:polynomial_examples} with different example datasets. 

The diffusion components converge to the eigenfunctions of the Fokker-Planck operator $\mathbf{H}$, if the data points are uniformly distributed on the manifold \cite{Nadler2007}. In the case of a non-intersecting 1-D curve, $\mathbf{H}\psi = \frac{d^2\psi}{dx^2}$, being $x$ the arc-length of the curve. If $\alpha=1$ or if the data points are uniformly distributed, the eigenfunctions are then given by 
\begin{equation}
\label{psi_n}
    \psi_n = cos(n\pi x)\text{,}
\end{equation} %if we solve the eigenvalue problem $H\psi = \lambda \psi$ by 
applying Neumann boundary conditions with $x=0,1$ at the edges \cite{Nadler2007}. In other words, $\dot{\Psi}(0)=\dot{\Psi}(1)=0$. 
$cos(nx)$ can be expressed for $n=1,\dots,5$ as:
\begin{align*}
    cos(x) &= 1\\
    cos(2x) &= -1 + 2cos^2(x)\\
    cos(3x) &= -3cos(x) + 4cos^3(x)\\
    cos(4x) &= 1-8cos^2(x)+8cos^4(x)\\
    cos(5x) &= 5cos(x)-20cos^3(x)+16cos^5(x)\\
    &...
\end{align*}
As $\psi_1 = cos(x)$ the larger diffusion components are polynomials of the first diffusion component, because they can be expressed as:
\begin{align}
\label{polynomials}
    \psi_2 &= -1 + 2\psi_1^2\\
    \psi_3 &= -3\psi_1 + 4\psi_1^3\\
    \psi_4 &= 1-8\psi_1^2+8\psi_1^4\\
    \psi_5 &= 5\psi_1-20\psi_1^3+16\psi_1^5\\
    &... \nonumber
\end{align}

\cite{Nadler2007} also mentions that if one assumes a closed one-dimensional manifold then one should use periodic boundary conditions. It follows that the dataset is mapped into a circle in the first two diffusion components.

\section{Diffusion Map: in practice}

\subsection{The Swiss roll example}
To illustrate the properties of the Diffusion Map and the influence of its parameters, we employ a classic synthetic dataset commonly used to demonstrate the effects of dimensionality-reduction techniques: the Swiss roll (for example in \cite{Tenenbaum2000,roweis2000, Nadler2007, Beier2025}).

\begin{figure}[]
    \centering
    \includegraphics[width=\linewidth]{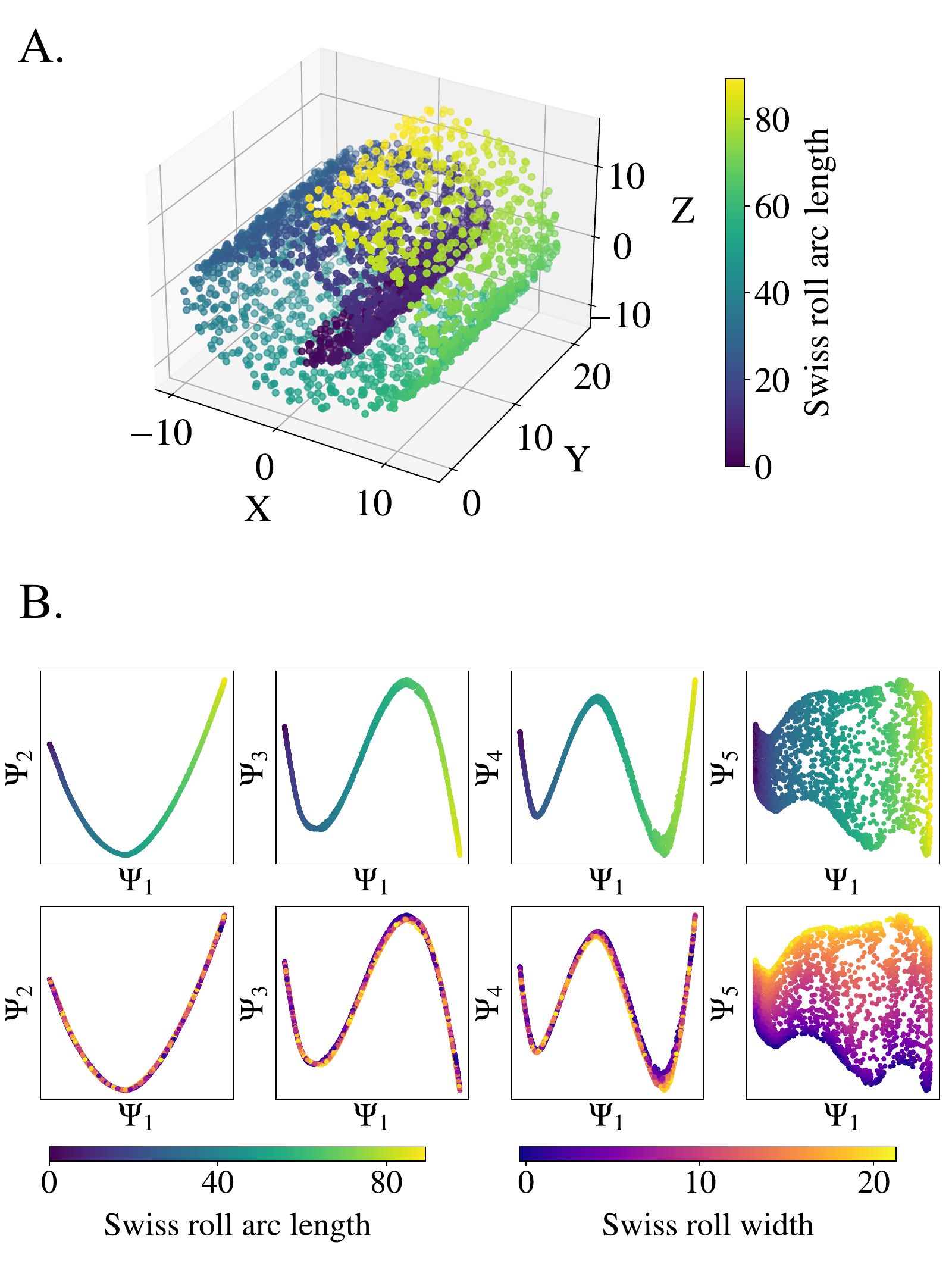}
    \caption{\textbf{The Swiss roll dataset}. \textbf{A.} shows the original dataset (a narrow Swiss roll) with parameters $(n,\sigma^2, H)=(3000, 0.2,21)$. The color encodes the length dimension of the sheet, which is rolled up to the Swiss roll (arc-length). The width of the Swiss roll is along the Y-axis. \textbf{B.} shows the Diffusion Map embedding considering the first 5 components to compare them. It is worth noting that all components are shown twice to visualize how both the arc length and the width are unfolded.
    The parametrization used was $(\epsilon, \alpha, t, N)=(5, 1/2, 1, 3000)$.} 
\label{fig:swiss_roll}
\end{figure}

This dataset consists of points sampled from a two-dimensional manifold that is nonlinearly embedded in three-dimensional Euclidean space. Geometrically, the dataset resembles a rolled-up sheet, where the intrinsic structure is essentially the arc length along the roll, with an additional width dimension (see Fig. \ref{fig:swiss_roll} A).

We use the implementation provided by \cite{scikit-learn}, where the Swiss roll is parameterized as follows:
\begin{equation}
    \label{eq:swissroll}
    \begin{pmatrix}
        x \\
        y \\
        z
    \end{pmatrix}
    =
    \begin{pmatrix}
        s \cos(s) + \xi_x \\
        h + \xi_z \\
        s \sin(s) + \xi_y \\
    \end{pmatrix},
\end{equation}
where $s$ and $h$ are sampled as $s \sim U([\frac{3 \pi}{2}, \frac{9 \pi}{2}])$ and $h \sim U([0, H])$.
We keep the width $H$ of the Swiss roll as a parameter. 
Also a small noise term $\xi_{x,y,z} \sim \mathcal{N}(0, \sigma^2)$ is included.

Because of its simple yet nonlinear geometry, the Swiss roll provides a controlled test case for assessing how well algorithms recover low-dimensional manifold structures from high-dimensional data. Linear methods such as PCA fail to capture the underlying geometry due to the nonlinear nature of the data set. Nonlinear techniques, including Diffusion Maps \cite{Nadler2005}, Isomap \cite{Tenenbaum2000}, or Locally Linear Embedding \cite{roweis2000}, can unfold the roll and reveal the intrinsic low-dimensional geometry.

In Figure \ref{fig:swiss_roll}B, we show a typical Diffusion Map embedding of a narrow Swiss roll, where the length of the rolled sheet is substantially greater than its width (cf. \cite{Nadler2007}). In this example, the width dimension appears only in the fifth diffusion component, whereas higher components represent polynomials of the first diffusion component, which encodes the arc length. This occurs because the width is much smaller in scale and therefore less important for the Diffusion Map algorithm. The emergence of such polynomial components is consistent with the behavior illustrated in Figure \ref{fig:polynomial_examples} for one-dimensional non-intersecting datasets, as the Swiss roll is effectively one dimensional with only a minor width dimension. In Section \ref{section_scaling}, we will demonstrate that the dataset’s two-dimensional structure emerges in the higher components when the natural parameters of the dataset share the same range and thus have equal importance.

In the following sections, we examine how parameter choices and dataset preprocessing influence the resulting embedding. 
While the Swiss roll dataset provides a clear and accessible test case, caution is required when generalizing observations to more complex or higher-dimensional datasets \cite{VanDerMaaten2009a,Strange2014}. As emphasized in \cite{Strange2014}, \textit{"Showing that an algorithm can perform well on the Swiss roll dataset does not offer any real insights into how the algorithm works or how the algorithm will work on real data.”} Nevertheless, we argue that if difficulties or artifacts already emerge when applying the method to a simple dataset, such as the Swiss roll, these issues are likely to persist or even become more pronounced when analyzing more complex real-world datasets. Accordingly, we use this dataset as a basis for an exploratory investigation of the method.

\subsection{Effect of the Diffusion Map parameters}
\subsubsection{Neighborhood width \texorpdfstring{$\epsilon$}{epsilon}}

\begin{figure}[]
\centering
\subfigure[Two first components, $\Psi_1$ and $\Psi_2$, of the Diffusion Map for different $\epsilon$ values of the Swiss roll with parameters $(n,\sigma^2, H)=(3000, 0.2,21)$. The color of the markers corresponds to the arc length. With $\epsilon = 5$ the manifold succeeds in unfolding the original data. The parametrization used was $(\alpha, t, N)=(1/2, 1, 3000)$.]{
   \centering
   \label{fig:dmap_epsilon_swissroll}
   \includegraphics[width=0.9\linewidth]{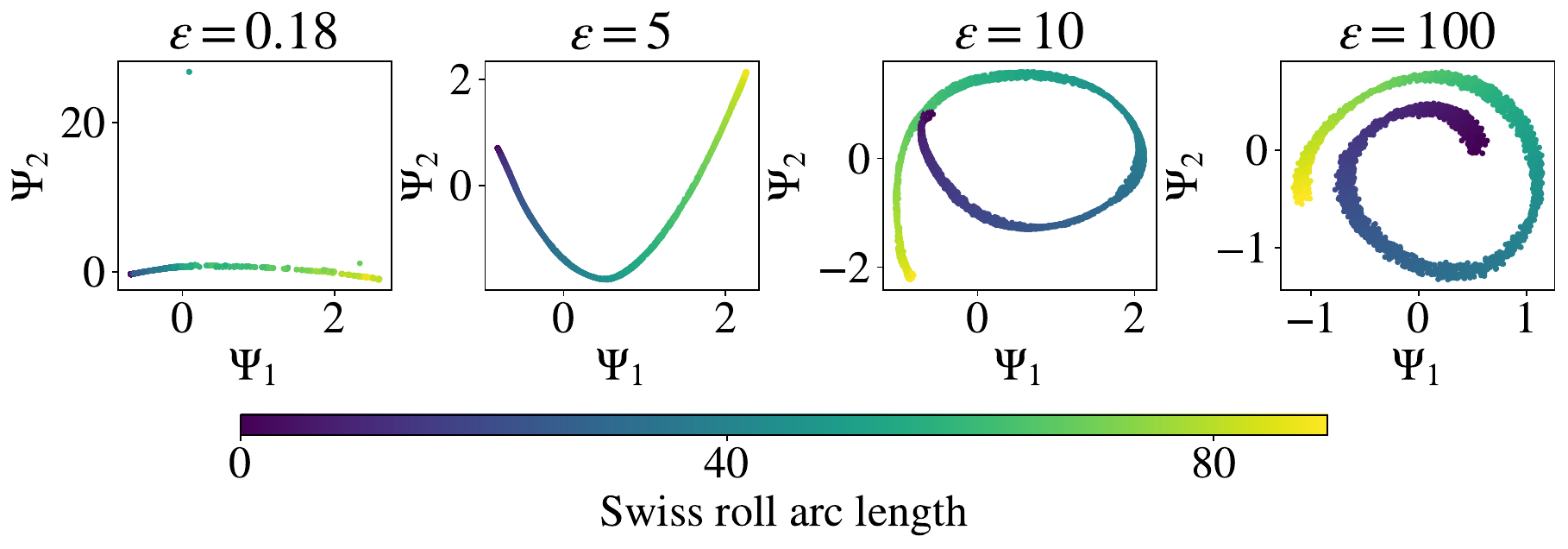}}~\\
\subfigure[Illustration of the effect of $\epsilon$ on the neighborhood defined by the kernel in the Diffusion Map construction. The plot shows the Swiss roll with parameters $(n,\sigma^2, H)=(3000, 0.2,21)$ color-coded according to the kernel applied on a certain data point $x_0$, $k(x_0,x_i)$, for $\epsilon=5$ (left) and $\epsilon=10$ (right).]{
   \centering 
   \label{fig:swissroll_epsilon_Ms}
   \includegraphics[width=0.7\linewidth]{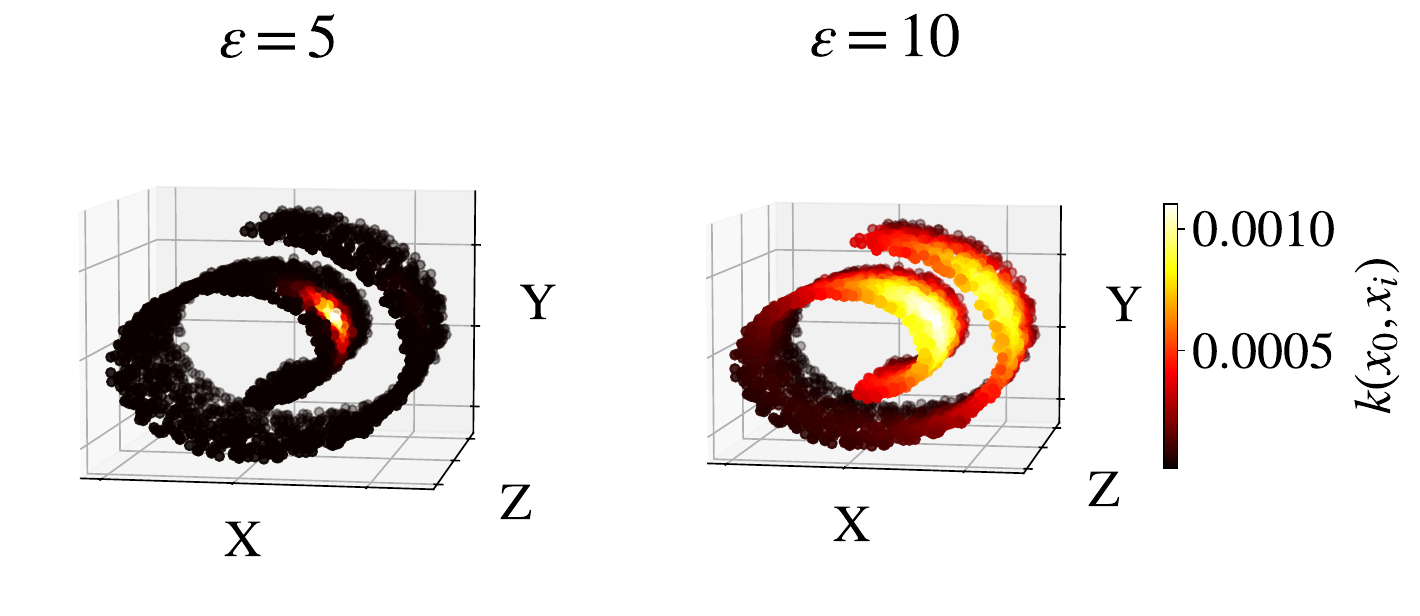}}~\\
\subfigure[Illustration of the equivalence of PCA and the Diffusion Map for a large enough parameter $\epsilon$. The Diffusion Map shows the same result as the PCA for $\epsilon = 1000$ and the rest of parameters were $(\alpha, t, N)=(1/2, 1, 3000)$. The color of the markers corresponds to the arc length, just as in Fig. \ref{fig:dmap_epsilon_swissroll}. The original data used is the same for both cases, Swiss roll with parameters $(n,\sigma^2, H)=(3000, 0.2,21)$.]{
   \centering
   \label{fig:PCA_diffmap_swissroll}
   \includegraphics[width=0.9\linewidth]{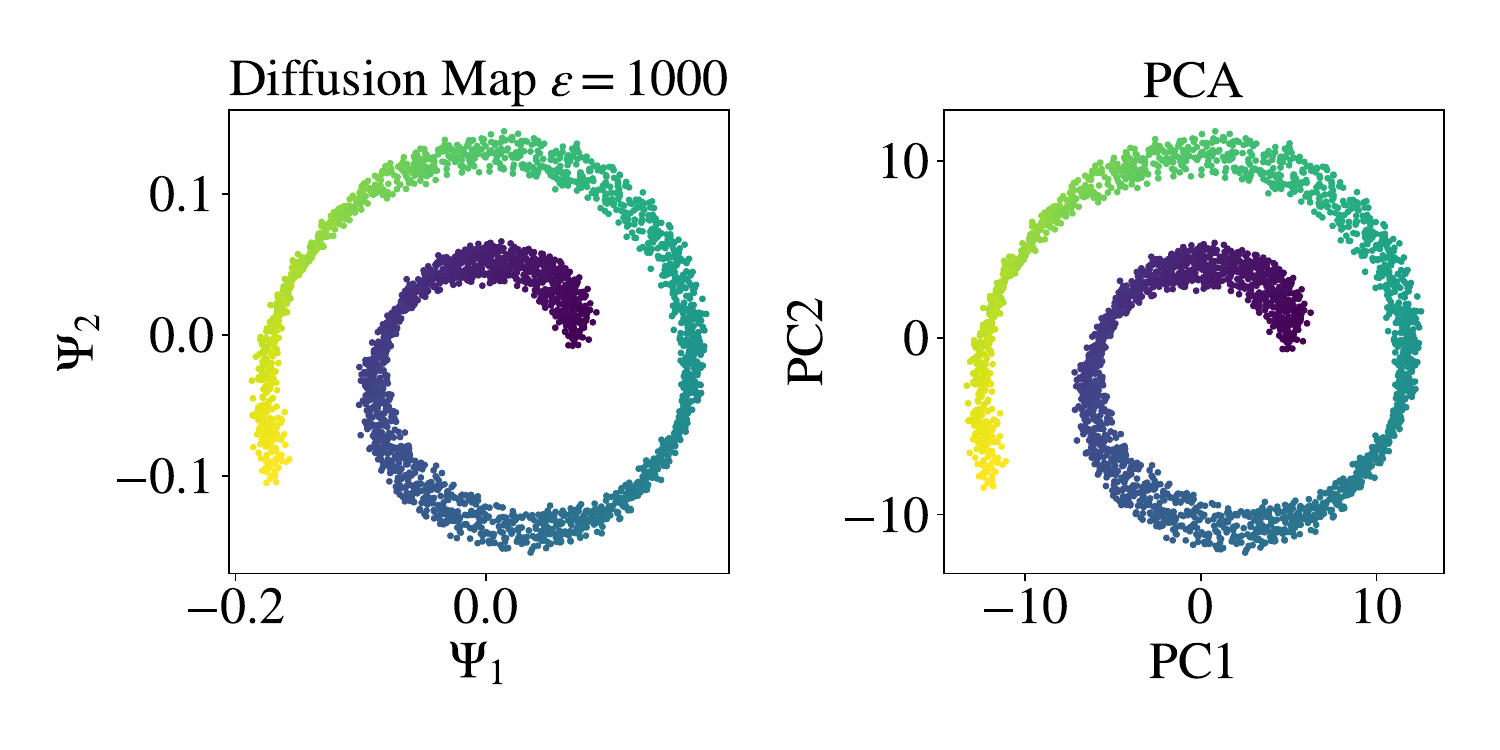}}
   \caption{\textbf{Visualization of the effect of the width parameter $\epsilon$} using the Swiss roll dataset.}
\end{figure}

%parameter role
The width kernel defined in Eq. \ref{kernel} plays a crucial role in the Diffusion Map framework, as it determines the size of the neighborhood in the graph construction.
%impact
If $\epsilon$ is chosen too small, the kernel captures only very close relationships, leading to a sparse or even disconnected graph. This situation leads to poor results, as the original data can no longer be represented on a well-interpretable manifold. For example, see case $\epsilon = 0.18$ in Figure \ref{fig:dmap_epsilon_swissroll}.
In contrast, if $\epsilon$ is chosen too large, the locality is lost and we obtain a highly-connected graph, where all data points exhibit high similarity. 

In Fig. \ref{fig:swissroll_epsilon_Ms} we illustrate the effect of a large $\epsilon$ in the transition probabilities. For $\epsilon=10$, jumps from one extreme of the manifold to a inner and locally disconnected part are allowed. When that happens, the Diffusion Map cannot follow the underlying manifold and cannot reproduce the intrinsic low-dimensional data (see Figure \ref{fig:dmap_epsilon_swissroll}). 

An appropriate choice of $\epsilon$ allows to capture the local geometric information, while keeping a coherent global representation (see $\epsilon = 5$ in Figure \ref{fig:dmap_epsilon_swissroll} and Figure \ref{fig:swissroll_epsilon_Ms}).
Consequently, $\epsilon$ determines the scale at which the manifold is analyzed and it is crucial for revealing its intrinsic low-dimensional structure. 

We observed that for a large enough $\epsilon$, the result of the Diffusion Map converges to that of PCA (see Fig. \ref{fig:PCA_diffmap_swissroll}). This behavior has also been observed for more complex datasets in \cite{Beier2025}. However, this is not surprising, since an excessively large neighborhood width removes locality, which is the main differing trait between Diffusion Maps and PCA: the consideration (or not) of locality. It is therefore advisable to always compute the PCA result and compare it with the outcome of the Diffusion Map. If the neighborhood is too large, no significant differences between the two representations can be observed and $\epsilon$ should be adjusted. \\

%discussion
An inappropriate choice of $\epsilon$ can substantially affect the outcome and may lead to incorrect conclusions. For example, \cite{Zhang2011} incorrectly reported that the Diffusion Map fails to unroll the Swiss roll—an observation that we suspect arises from an improper choice of neighborhood size.\\
The correct value of $\epsilon$ depends on various factors, such as the number of data points, the dimension, the manifold volume or the manifold curvature \cite{Singer2006,Meila2024}. \\
Although several approaches for determining a suitable value of $\epsilon$ have been proposed, for example, using the minimum squared distance \cite{lafon2004,Cameron2021} or the sum of the entries of the kernel matrix \cite{Singer2009,Bah2008}, these methods generally provide only an initial estimate. Indeed, there are examples where the proposed techniques fail to yield an appropriate choice of $\epsilon$ \cite{Beier2025}.\\

\subsubsection{Number of nearest neighbors considered \texorpdfstring{$N$}{N}}

\begin{figure}[]
    \centering
    \includegraphics[width=\linewidth]{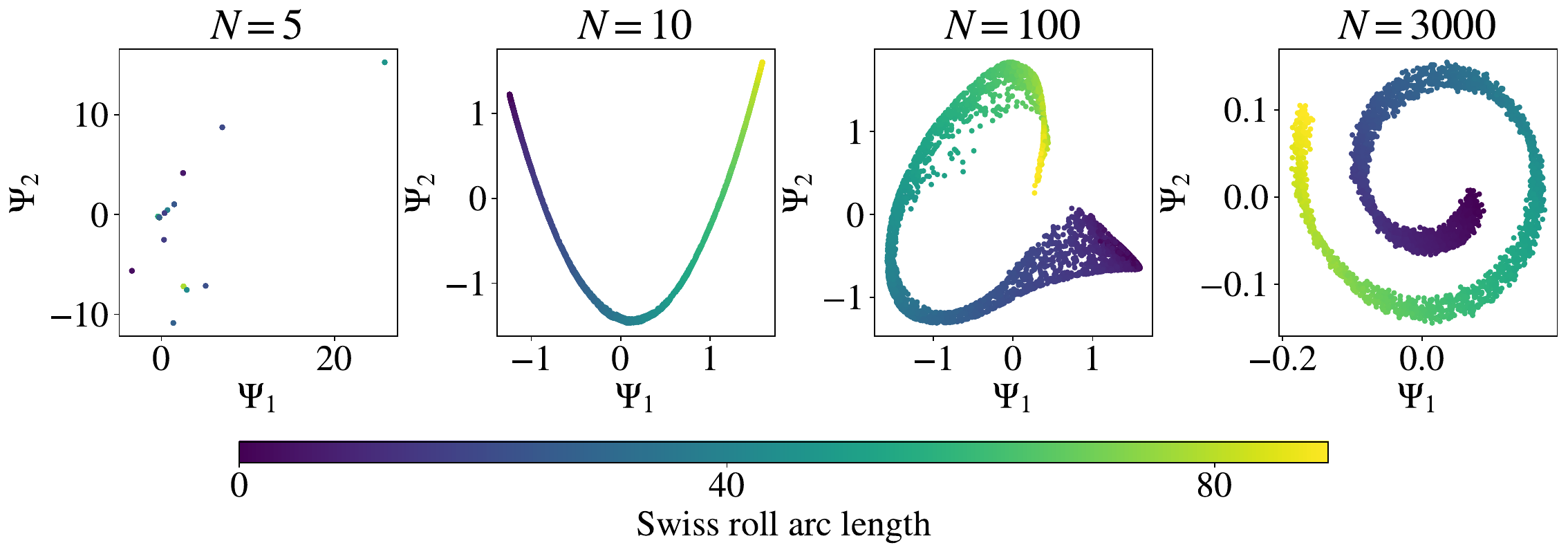}
    \caption{\textbf{Illustration of the effect of the number of considered nearest neighbors $N$} using the Swiss roll with parameters $(n,\sigma^2, H)=(3000, 0.2,21)$. $(\epsilon, \alpha, t) =(1000,1/2,1)$ were used as Diffusion Map parameters.} 
    \label{fig:N_swissroll}
\end{figure}

%parameter role
Another way to define locality is by limiting the considered number of nearest neighbors $N$. This method is used, on the one hand, because it turns the matrices considered into sparse matrices, which are computationally more efficient to handle \cite{DeSilva2002,Luxburg2007,VanDerMaaten2009a,Barter2019,Meila2024}. 
Some authors recommend this N-nearest neighbors approach, because it is easier to choose a suitable parameter \cite{Luxburg2007}.\\
%impact
What has received less attention so far is that when both parameters are used, $N$ not only reduces computational cost but also interacts with $\epsilon$ in shaping the Diffusion Map embedding. As shown in Figure \ref{fig:N_swissroll}, even if $\epsilon = 1000$ is too large to properly unfold the manifold, a small $N$ (e.g., $N = 10$) can still recover the expected result, suggesting that $N$ may act as the dominant neighborhood parameter. Setting $N$ too small leads to a disconnected graph (see, $N=4$). 
%illustration

\subsubsection{Time parameter \texorpdfstring{$t$}{t}}
\begin{figure}[]
    \centering
    \includegraphics[width=\linewidth]{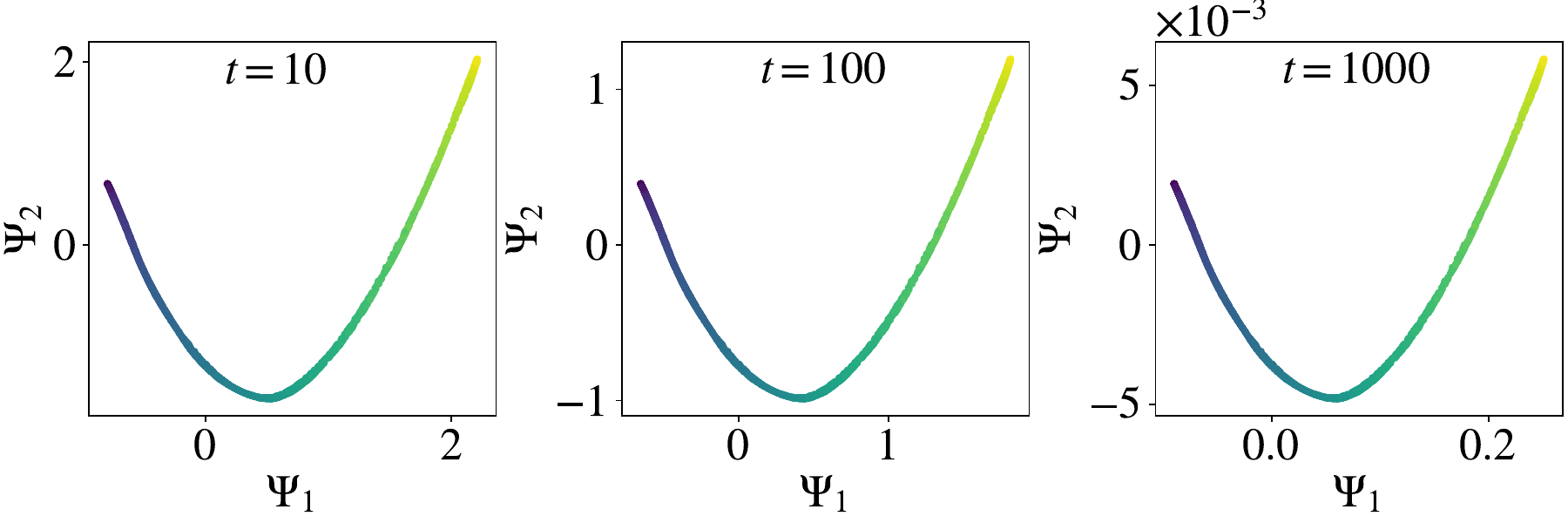}
    \caption{\textbf{Illustrating the effect of the time parameter $t$} using the Swiss roll with parameters $(n,\sigma^2, H)=(3000, 0.2,21)$. $(\epsilon, \alpha, N) =(5,1/2,3000)$ were used as Diffusion Map parameters.}
    \label{fig:influence_t_to_swissroll}
\end{figure}

%parameter role
The parameter $t$ was originally introduced as a time parameter \cite{lafon2004,Nadler2005,Coifman2005}, with the transition matrix $M$ raised to the power $t$ to represent $t$ steps of the associated Markov process. In later works, it is stated that $t$ enables the exploration of the geometric structure of the original dataset at different scales and should therefore be considered a scale parameter \cite{Coifman2006,Porte2008}. According to \cite{Coifman2006}, this temporal evolution is also a key idea of the Diffusion Map framework, distinguishing it from the Laplacian Eigenmap defined in \cite{Belkin2001} and highlighting the role of diffusion along the manifold in producing meaningful embeddings. Furthermore, $t$ should contribute to robustness against noise, as increasing $t$ averages over all possible $t$-step paths, thereby smoothing local perturbations \cite{lafon2004,Coifman2006}.\\
%impact & illustration
We present an opposing view, suggesting that the parameter $t$ is often unnecessary and offers limited benefits for data analysis. 
As argued in \cite{Saul2003} the spectral decomposition already links the local geometric information to reveal the global structure therefore letting the Markov chain evolve is not necessary. Often $t=1$ is used without justification. For $t=0$ the definition of the Diffusion Map components becomes equivalent to the definition of the Laplacian eigenmap components \cite{Belkin2003}.
%The Markov chain is not required to capture the manifold. As argued in \cite{Saul2003} the spectral decomposition without using $t$ already links local geometric information to reveal the global structure. 
Many other spectral methods, for example kernel PCA \cite{schoelkopf1997}, Local Linear Embedding \cite{roweis2000}, Spectral Clustering/Embedding \cite{Ng2001} and the named Laplacian Eigenmaps method \cite{Belkin2001}, similarly do not have a $t$ parameter and still achieve meaningful and reliable results.\\
To illustrate this, Figure \ref{fig:influence_t_to_swissroll} shows that varying $t$ does not alter the structure of the Diffusion Map for the Swiss roll. Only the axis, i.e. the diffusion components $\Psi_1$,$\Psi_2$ are rescaled to different extents. With increasing $t$ the Diffusion Map shrinks, but its structure remains unchanged. This effect can also be observed in more complex datasets, such as the political science data analyzed in \cite{PirkerDiaz2025,Beier2025}.\\
%discussion
This behavior can be explained by looking at the definition of the Diffusion Map: the $i$-th diffusion component is given by $\lambda_i^t \Psi_i(x)$, where $\lambda_i$ and $\Psi_i$ denote the eigenvalues and eigenvectors of $M_s$, respectively, and are independent of $t$. 

Thus, the only influence on the Diffusion Map is the exponentiation of the eigenvalues. Since $\lambda_i < 1$, the eigenvectors are merely scaled by $\lambda_i^t$, so larger $t$ values decrease the diffusion components at different rates, with higher components decaying more rapidly. Although this affects the distances within the embedding, the overall structure remains unchanged. Although the effect of the $t$ parameter could be important from a theoretical perspective, it is of little use for qualitative analysis of high-dimensional data.

\subsubsection{Anisotropic kernel parameter \texorpdfstring{$\alpha$}{alpha}}\label{alphasection}

%parameter role
The anisotropic kernel parameter $\alpha$ adjusts the amount of influence of the data density on the transitions of the diffusion process \cite{Coifman2006}, and consequently, on the resulting embedding.\\
%illustration
% Figure \ref{fig:swissroll_alpha_comparison} illustrates its effect by showing two Diffusion Map embeddings of the same Swiss roll (see Fig. \ref{fig:swiss_roll}A) but with a different choice of $\alpha$.
% It can be seen that in both cases, points from the inner end of the Swiss roll (blue) are closer together than the ones of the other end (yellow).
% This is because the densities in both extremes differ in the original data.

For $\alpha = 0$, this effect is not compensated and the influence of the density is maximal, leading to a greater density fluctuation across the manifold. %That is something that we can observe in Fig. \ref{fig:swissroll_alpha_comparison} (left), where we can see large pockets containing no points, which are the result of the exaggeration of such fluctuations, and a greater density fluctuation across the manifold.

On the other hand, for $\alpha=1$, the kernel is normalized such that the effects of sampling density are reduced, as mentioned in section \ref{asymptoticssection}. \\ %This effect can be seen in Fig. \ref{fig:swissroll_alpha_comparison} (right). \\

%discussion
In conclusion, $\alpha$ balances locality and geometry. By tuning it, we choose to be more sensible to the data density ($\alpha\to0$) or to the geometry ($\alpha\to1$), specially when working with non-uniformly sampled data. 
It is worth mentioning that there is no noticeably difference in the computational cost of calculating the Diffusion Map with different $\alpha$ values.

% \begin{figure}
%     \centering
%     \includegraphics[width=\linewidth]{figures2/fig7.pdf}
%     \caption{\textbf{Showing the effect of the anisotropic kernel parameter $\alpha$} Diffusion Maps of the Swiss roll with parameters $(n,\sigma^2, H)=(2000, 0.2,21)$ under different choices of $\alpha$, 0 (left) and 1 (right). $(\epsilon, N, t) =(1,100,1)$ were used as Diffusion Map parameters.}
%     \label{fig:swissroll_alpha_comparison}
% \end{figure}

% \begin{figure}[htpb]
%     \subfigure[$\alpha = 0$]{
%         \centering
%         \label{fig:swissroll_dm_alpha_0}
%         \includegraphics[width=\linewidth]{swissroll_dm_eps_4_alpha_0.png}
%         }~\\
%     \subfigure[$\alpha = 1$]{
%         \centering
%         \label{fig:swissroll_dm_alpha_1}
%         \includegraphics[width=\linewidth]{swissroll_dm_eps_4_alpha_1.png}
%         }~\\
%     \caption{Diffusion Maps of the Swiss roll under different choices of $\alpha$. In both cases, $\epsilon=4$ was used. The original data corresponds to the one plotted in Fig. \ref{fig:swiss_roll}A.}
%     \label{fig:swissroll_alpha_comparison}
% \end{figure}

\subsection{Effect of the data preprocessing}
In preparing data for the application of the Diffusion Map, various preprocessing techniques are often employed. Among them, individual variables may be scaled, normalized or standardized (for example in \cite{Barter2019,PirkerDiaz2025,Zeng2024}). When collecting data, it may also occur that some information is redundantly represented across multiple variables and statistics, as is often the case, for example, in census datasets \cite{Barter2019,Beier2025}. In addition, variables can be continuous or discrete, what can have a big impact on the resulting Diffusion Map \cite{Beier2025}. 
To date, these issues have received limited consideration in practical applications. Therefore, we examine in this section how preprocessing of data can influence the outcome of the Diffusion Map.

\subsubsection{Scaling}
\label{section_scaling}
Transformations such as rescaling certain variables can have a substantial impact on the outcome of the Diffusion Map algorithm. This effect was also indicated in \cite{Nadler2007}, where altering the width dimension of the Swiss roll dataset led to markedly different results, yielding either a 1D perspective of the dataset—capturing only the arc-length dimension or a 2D perspective, in which the width of the Swiss roll is also reflected in the Diffusion Map manifold.\\

\begin{figure}
    \centering
    \includegraphics[width=\linewidth]{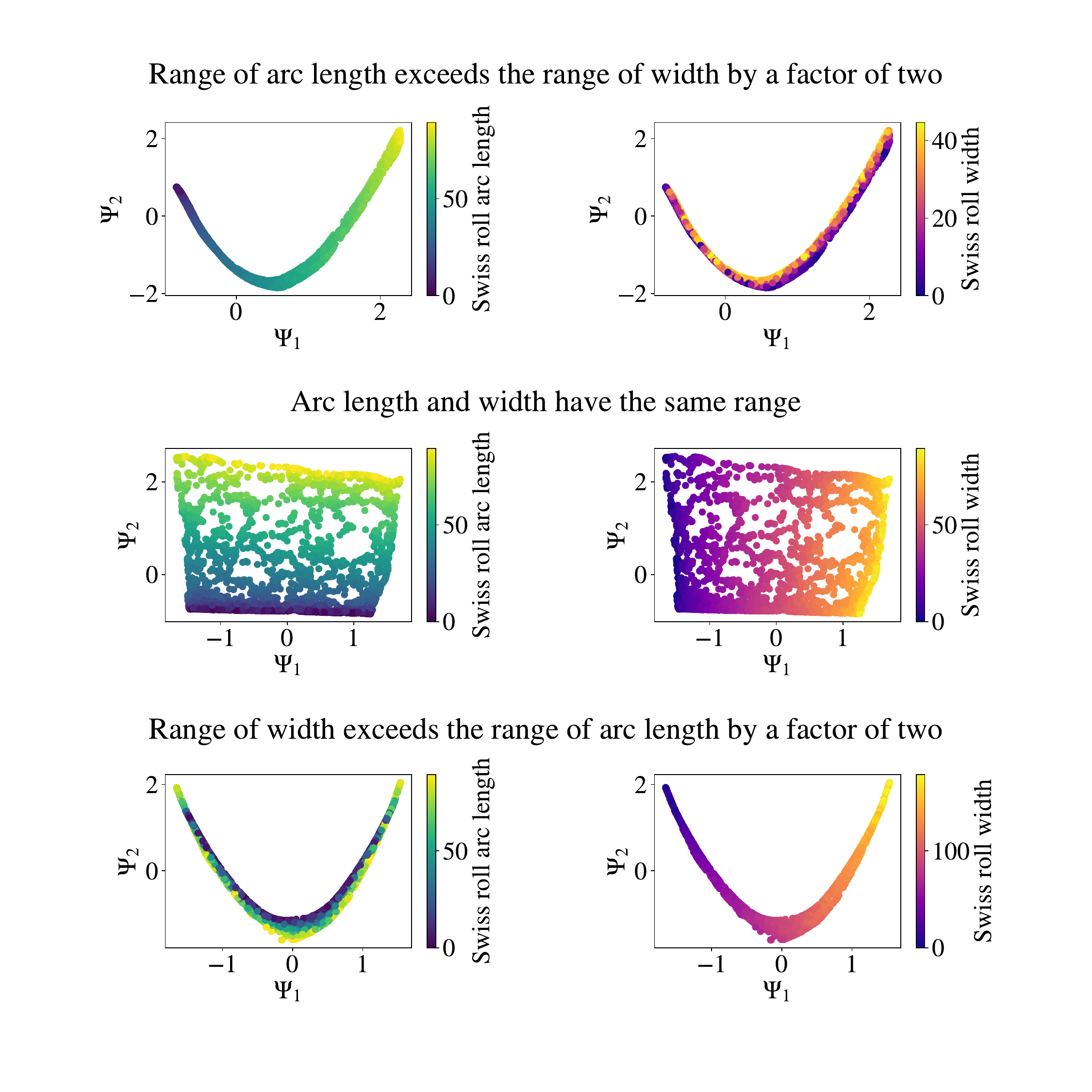}
    \caption{\textbf{Showing the effect of rescaling the $Y$-axis} and therefore the width of the Swiss roll. Each row corresponds to a different rescaling of the original data, the Swiss roll with parameters $(n,\sigma^2,H)=(3000,0.2,21)$. For each row, the the color code corresponds to the arc length along the Swiss roll (left) and to the width of the Swiss roll (right). $(N, t, \alpha, \epsilon)=(3000, 1, 1/2,5)$ was used for all datasets. The first row of figures corresponds to a narrow Swiss roll. The second row corresponds to a wide Swiss roll, where the shape of the folded sheet is squared.}
    \label{swiss_roll_manipulate_dimension}
\end{figure}

Here, we reproduce this effect and further investigate its implications and significance for the application of the Diffusion Map. To do so, we examine the Diffusion Map of the Swiss roll in Figure \ref{swiss_roll_manipulate_dimension}, where we systematically vary the width dimension and analyze how both the arc length and the width of the dataset are represented in the embedding. This allows us also to determine under which conditions the Diffusion Map yields either a 1D- or 2D-representation.

If we scale the width dimension such that its range is five times smaller than the range of the arc-length dimension, the Diffusion Map produces a one-dimensional embedding in which the arc length is represented by $\Psi_1$, consistent with the results in Figure \ref{fig:dmap_epsilon_swissroll} and \cite{Nadler2005}, where the width dimension is also smaller.

When the width of the Swiss roll is scaled such that it has the same range as the arc-length dimension, the resulting embedding becomes two-dimensional, with the first two diffusion components $\Psi_1$ and $\Psi_2$ encoding the arc length and the width, respectively.

Increasing the scaling further such that the range of the width becomes five times larger than the range of the arc-length again leads to a one-dimensional representation. However, in this case the arc-length information disappears from the leading components, and the first diffusion component $\Psi_1$ now reflects the width.\\

In summary, scaling can substantially alter the relative importance of the original variables for the Diffusion Map algorithm. This is not surprising, as the method is based on pairwise distances between data points, and enlarging a single dimension automatically increases its contribution to these distances. Nevertheless, this effect must be taken into account when designing new datasets and applying preprocessing transformations to datasets. Otherwise, one risks drawing incorrect conclusions from the Diffusion Map by mistakenly attributing greater intrinsic importance to a rescaled dimension, when in fact this importance is merely an artifact of the transformation.\\

It is also worth noting that scaling leads to different optimal values for the Diffusion Map parameters, which must be adjusted accordingly.

\subsubsection{Normalization}
%The comparison of \textcolor{red}{Figure \ref{swiss_roll_normalization} with Figure \ref{fig:dmap_epsilon_swissroll}} shows that different normalizations of the Swiss roll dataset lead to different optimal values of $\epsilon$. 
%Different normalizations of the Swiss roll dataset lead to different optimal values of $\epsilon$. For instance, when standardizing the data, i.e. that each variable is adjusted to have a mean of zero and a standard deviation of one, (like in figure \ref{fig:dmap_epsilon_swissroll}) an optimal $\epsilon$ of around 0.1 yields the expected parabolic shape. Without standardization, using the original scale of the data, the optimal value shifts to approximately $\epsilon = 1$, whereas rescaling all variables to the interval $[0, 1]$ requires an $\epsilon$ of about $0.01$.

%This is expected, as normalization alters pairwise distances. Since the Diffusion Map algorithm fundamentally relies on these distances, the optimal neighborhood parameter $\epsilon$ changes accordingly. 

Datasets are often normalized prior to analysis, for example through standardization or min–max normalization. These procedures can be understood as special cases of dataset scaling. It is important to note that such preprocessing steps can also have a substantial impact on the resulting embedding.\\

Using the Swiss roll as an example, scaling the $x$,$y$, and $z$ coordinates to the same range for example by min-max normalization implies that the arc-length dimension becomes larger than the width dimensions. As shown in the previous sections, this leads to an almost one-dimensional diffusion map. Other scaling factors can lead to different results, as we saw in the previous chapter.\\

When deciding on a normalization, one should consider the meaning and units of the different dimensions in the original dataset.

For instance, if all or some of the variables represent different axes in some Euclidean space, it likely does not make sense to normalize them, since the relative scaling of the variables does have meaning.

However, if the variables represent measurements of various quantities which are measured in different units, then there is no ground truth as to the relative scaling and normalization may be helpful to enable using Diffusion Map. One should keep in mind that creating new variables with specific normalizations—for example in social-science datasets such as census data—may inadvertently alter the diffusion map outcome.

% \begin{figure}
%     \centering
%     \includegraphics[width=\linewidth]{figures2/fig8.pdf}
%     \caption{\textbf{Showing the effect of standardization and rescaling} for the Diffusion Map embedding.    
%     Considering the same original dataset, the Swiss roll with parameters $(n,\sigma^2, H)=(2000, 0.2,21)$, it is evident that the selection of $\epsilon$ must be adapted in order to achieve comparable results, depending on the presence or absence of data standardization. In both cases, $(\alpha, N, t) =(1/2,100,1)$ were used as Diffusion Map parameters}
%     \label{swiss_roll_normalization}
% \end{figure}

\subsubsection{Discrete and continuous variables}
When the Diffusion Map is applied to various real-world datasets, such as social science census data or democracy indicators, it is often the case that some variables are continuous while others are discrete or quasidiscrete.

Consider, for example, the V-Dem democracy dataset comprising all countries and years since 1900 for different aspects of democracy, as examined in \cite{Beier2025}. Variables such as media freedom, which are based on expert assessments, yield relatively continuous measures (cf. \cite{Beier2025}, Figure 4.5). In contrast, the suffrage indicator—measured as the percentage of the population with voting rights—constitutes a discrete variable, since in practice only three values commonly occur: 1 (universal suffrage), 0.5 (exclusion of women), or very small values (reflecting the exclusion of large population groups, e.g., under racism or apartheid).

This characteristic has a significant impact on the outcome of the Diffusion Map: suffrage has a significant influence on the Diffusion Map embedding of the V-Dem dataset \cite{Beier2025}. This can be explained by the fact that pairwise distances between data points are strongly affected by the relatively large distances between data points within the discrete variables.\\
% We illustrate this effect in Figure \ref{fig:swiss_roll_discrete_variables} using the Swiss roll dataset. \textcolor{red}{Here, the width dimension of the original dataset (Figure \ref{fig:swiss_roll}) is replaced so that it no longer varies continuously between 0 and 4, but instead only takes the values $0$ and $\epsilon = 0.1$. As a result, the Diffusion Map highlights this discrete variable in its leading diffusion components, which shows the enhanced influence of this variable on the algorithm.} \\

Hence, when working with datasets containing both discrete and continuous variables, one must carefully consider whether certain discrete variables should be excluded prior to the analysis.

% \begin{figure}
%     \centering
%     \includegraphics[width=\linewidth]{figures/swiss_roll_discrete_width_diffusionmap.png}
%     \caption{\textbf{Illustrating the effect of discrete variables:} The plot shows the Diffusion Map result for $\epsilon = 0.1$ for the Swiss roll dataset in Figure \ref{fig:swiss_roll}, where the continuous width dimension reaching from $0$ to $4$ is replaced by a discrete variable with only the values $0$ and $\epsilon$.}
%     \label{fig:swiss_roll_discrete_variables}
% \end{figure}

\subsubsection{Data redundancy}

When analyzing datasets with the Diffusion Map, it is important to consider the intrinsic characteristics of the data itself. Most theoretical considerations have focused on data in which the variables are relatively independent and of roughly equal importance with respect to the purpose of the dataset.

However, this assumption may not hold in practice, as certain variables can be highly correlated, encode the same information, or originate from the same underlying construct.
In these cases there is an unequal contribution, as some variables are to some extent redundant. 

The Diffusion Map method does not distinguish any type of redundancy among variables in the data. Consequently, when applied directly to all variables -without previously identifying and filtering redundant ones- it treats each variable equally. 

It is therefore necessary to investigate the presence and influence of redundant or highly correlated variables before applying the Diffusion Map. Otherwise, careless application may produce unintended biased outcomes.

\begin{figure}
    \centering
    \includegraphics[width=\linewidth]{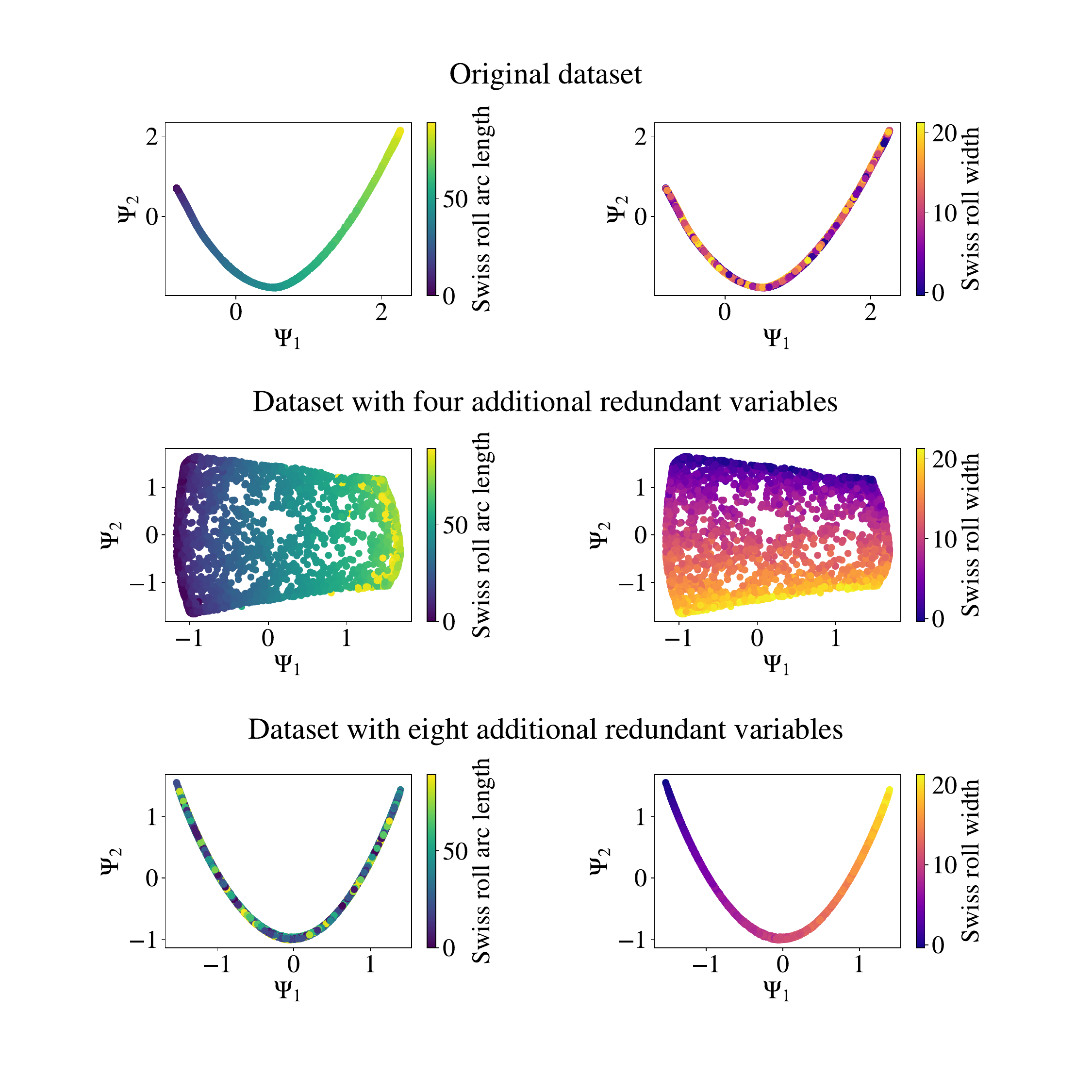}
    \caption{\textbf{Illustrating possible effects of redundant variables:} 
    The original dataset is a Swiss roll, created by using the \texttt{make\_swiss\_roll} method from \cite{scikit-learn} with the parameters \texttt{n\_samples=3000, noise=0.01}, to which further redundant dimensions/variables have been added. These added dimensions are copies of the width dimension of the Swiss roll (which is the Y-dimension in Figure \ref{fig:swiss_roll}) and Gaussian noise with a standard deviation of $1/10$ of the Swiss roll width was added. The duplicated (width-) dimension appears to have an increased influence on the Diffusion Map result. This effect becomes more pronounced as the number of redundant dimensions in the dataset increases.}
    \label{swiss_roll_redundant_variables}
\end{figure}

To illustrate this effect we refer again to our test dataset, the Swiss roll, shown in Figure \ref{fig:swiss_roll}. To investigate the influence of potential redundancies, we duplicate one of the original dimensions. Figure \ref{swiss_roll_redundant_variables} shows the diffusion map embeddings for the first diffusion components of the original dataset, as well as for datasets in which the width dimension $y$ is duplicated four and eight times, resulting in datasets of dimensionality seven and eleven.

Without redundant variables, the leading diffusion components capture only the arc-length information of the data points along the Swiss roll. As the number of redundant variables increases, however, the width dimension gains importance from the perspective of the method. With four redundant variables, the width dimension is encoded by the second diffusion component $\Psi_2$ and the two-dimensional nature of the diffusion map becomes apparent. With ten redundant variables, the arc-length information disappears entirely, and the first two components reflect only the width dimension.

This result is reminiscent of the behavior observed when scaling variables (see Figure \ref{swiss_roll_manipulate_dimension}), where rescaling the width dimension likewise produced the same effect.

%The reason for this behavior is that repeating a variable $q$ 
%times is equivalent to scaling it by $\sqrt{q}$. 
%Dies kann man anhand des Satzes von Pythagoras nachvollziehen: 

In conclusion, the Diffusion Map thus assigns greater importance to information represented by multiple redundant variables, since distances along this dimension gain weight. This effect must be considered in analyses using the Diffusion Map.

\subsection{Measuring the importance of Diffusion Map components}\label{sectioncomponents}

Determining how many diffusion components to retain is a crucial step for effectively reducing the dimensionality of the data.
When constructing the Diffusion Map, retaining only the first $k$ components is the standard way to proceed \cite{Coifman2006}.
However, it is not clear how $k$ should be chosen.
In practice, it is tempting to choose $k=2$, because that is the number of components that can be easily visualized.
Of course, this is not a sufficient justification; in the following, we shall discuss how to determine $k$ in order to faithfully represent the data.

Since the equivalence of Euclidean distance in the embedded space and \textit{diffusion distance} in the data space is the distinguishing feature of the Diffusion Map, it is natural to turn to this property to evaluate the accuracy of a given mapping.
This is why evaluating the spectrum $\lambda^t$ is the standard approach to evaluate the accuracy of the Diffusion Map in the literature.
% While this might be suited to analyze the diffusion process on a given timescale $t$, in practice, it is hardly helpful for the purposes of dimensionality reduction.
However, the spectrum $\lambda^t$ is highly dependent on $t$, as noted e.g.\ in \cite{Coifman2006}.
While this makes sense under the probabilistic interpretation of the Diffusion Map, for the purposes of dimensionality reduction, this $t$-dependence is problematic.
After all, any choice of $k$ could be justified by just appropriately tuning the time parameter $t$.
\begin{figure}
    \centering
    \includegraphics[width=\linewidth]{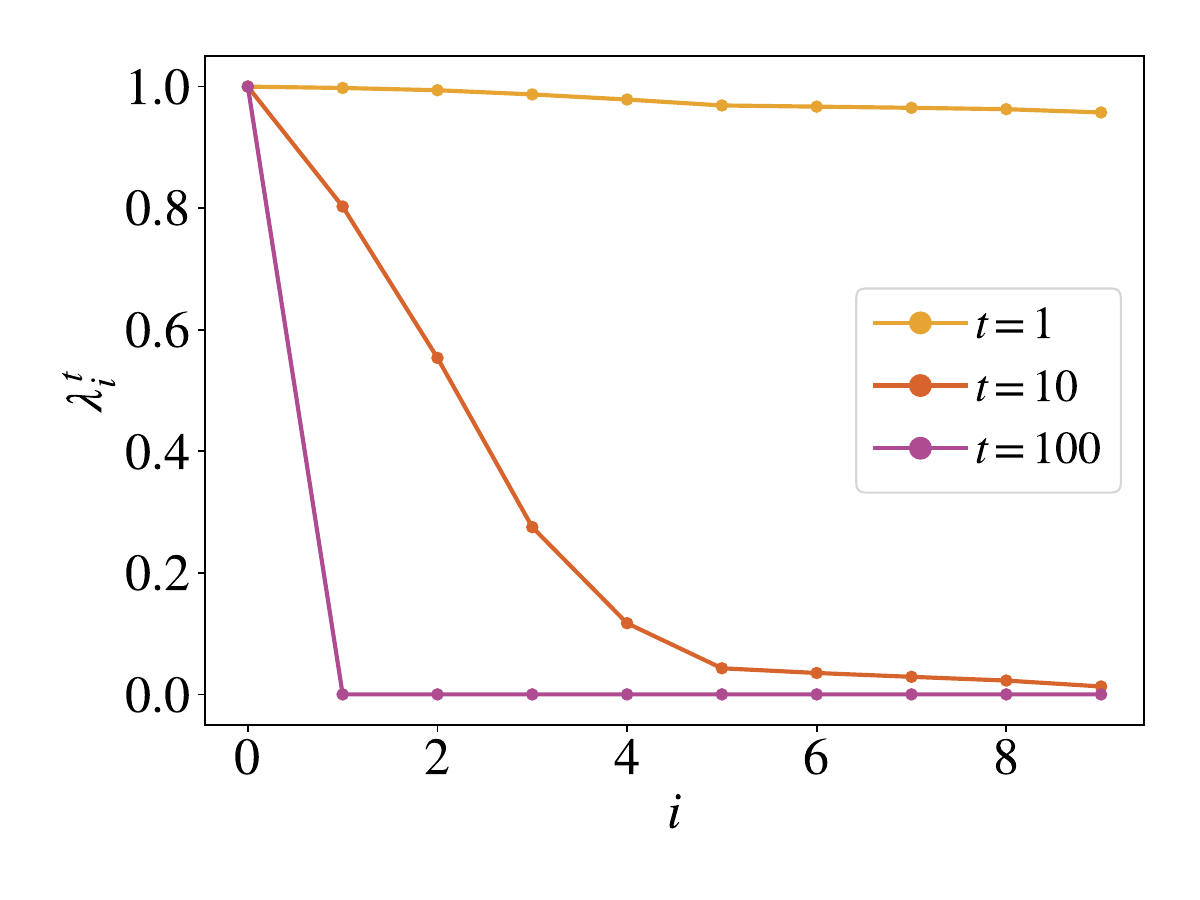}
    \caption{\textbf{Eigenvalue spectrum $\lambda^t$} of the Diffusion Map of the Swiss roll with parameters $(n,\sigma^2, H)=(2000, 0.2,21)$ with different time parameters $t$. In all cases, $(\alpha, N, \epsilon) =(1/2,700,1)$ were used as Diffusion Map parameters.}
    \label{fig:swissroll_spectrum}
\end{figure}

To illustrate this, Figure \ref{fig:swissroll_spectrum} shows the spectrum of the Diffusion Map of the Swiss roll dataset.
Because the decay is very shallow for low $t$, hundreds of diffusion components can remain above the 0.1 threshold --- a value commonly used to indicate a relevant precision level \cite{Coifman2005, Coifman2005Geometric}. For high $t$, the decay can be so sharp that only one component lies above that threshold.
In no case is the two-dimensional nature of the Swiss roll manifold revealed; there is no kink or any indication that distinguishes the first two diffusion components from the rest.
To fill the methodological gaps discussed above, in the following section, we showcase a method for identifying a suitable choice of components \cite{Pagenkopf2025, pagenkopf2026quantifying}. 

\subsubsection{The Neural Reconstruction Error}
The Neural Reconstruction Error (NRE) was developed recently to evaluate the accuracy of a Diffusion Map \cite{Pagenkopf2025}.
The method quantifies the importance of individual diffusion components and, thus, supports an accurate interpretation of the results of the Diffusion Map.

% Let $\left\{x_i \right\}_{i=1}^N$ be a set of data points ($x_i \in \mathbb{R}^p$) and $\Psi^{(k)}(x) \in \mathbb{R}^k$ be the Diffusion Map of $x$ using $k$ components.
The NRE method is based on the following argument. If $\hat{\Psi}^{-1}$ was an approximate inverse mapping of the Diffusion Map $\Psi$ (equation \ref{eq:diffusion_map}) such that $\hat{\Psi}^{-1}(\Psi(x)) \approx x$, then a straightforward measure of the accuracy of $\Psi$ would be the mean squared \emph{reconstruction error}:
\begin{equation}
    \label{eq:re}
    \begin{aligned}
        \varepsilon_k &= \frac{1}{p} \left\langle ||\hat{\Psi}^{-1}(\Psi(x)) - x||^2 \right\rangle \\
        &= \frac{1}{Np} \sum_{i=1}^{N} ||\hat{\Psi}^{-1}(\Psi(x_i)) - x_i||^2 ,
    \end{aligned}
\end{equation}
which is a function of the number $k$ of included diffusion components.
The averaging $\left\langle \cdot \right\rangle $ refers to averaging over all the points in the dataset.
For parametric dimensionality reduction methods for which this inverse mapping exists, this measure is straightforward to use.
Thanks to its straightforward and intuitive definition, it is considered an obvious and fair quality measure \cite{Lee2009, pyDR}.

Unfortunately, Diffusion Map, just like most other nonlinear dimensionality reduction methods, does not provide such an inverse mapping \cite{Lee2009}.
In \cite{Pagenkopf2025, pagenkopf2026quantifying}, the authors therefore turned to deep learning and employed a neural network to learn an approximate inverse mapping from the data.
The network receives the Diffusion Map coordinates $\{\Psi(x_i)\}$ as inputs and original data points $\{x_i\}$ as targets.
The reconstruction error is used as the cost function, such that the network will aim to minimize it.
After training, the reconstruction error (\ref{eq:re}) can be evaluated on the dataset using the approximate inverse mapping learned by the network.
Repeating this procedure for the different possible values of $k$ yields the $k$-dependent reconstruction error $\varepsilon_k$, called \emph{Neural Reconstruction Error (NRE)}.
% We will show how it can be used to identify relevant diffusion components and to help interpret the Diffusion Map.

%\begin{figure}[!ht]
%\centering
%\def \globalscale {0.800000}
%\input{nre_scheme.tex}
%\caption{Schematic illustration of NRE analysis}
%\label{fig:nre_scheme}
%\end{figure}

\begin{figure}[!ht]
\centering
\includegraphics[width=\linewidth]{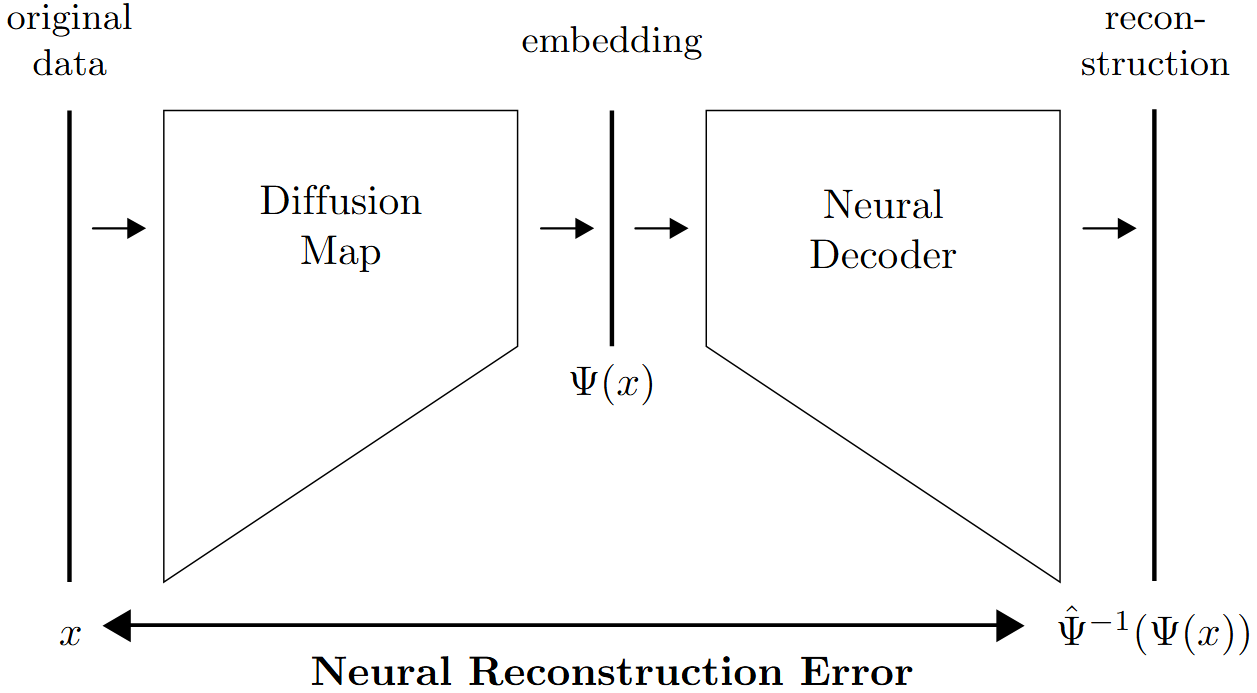}
\caption{Schematic illustration of NRE analysis}
\label{fig:nre_scheme}
\end{figure}

Figure \ref{fig:nre_scheme} illustrates the NRE schematically.
The structure is identical to an autoencoder, except that the encoder has been replaced with Diffusion Map. For details of the method, see the Appendix and \cite{Pagenkopf2025}.

%\subsubsection{Illustration of the NRE}

\begin{figure}[htpb]
    \centering
    \includegraphics[width=\linewidth]{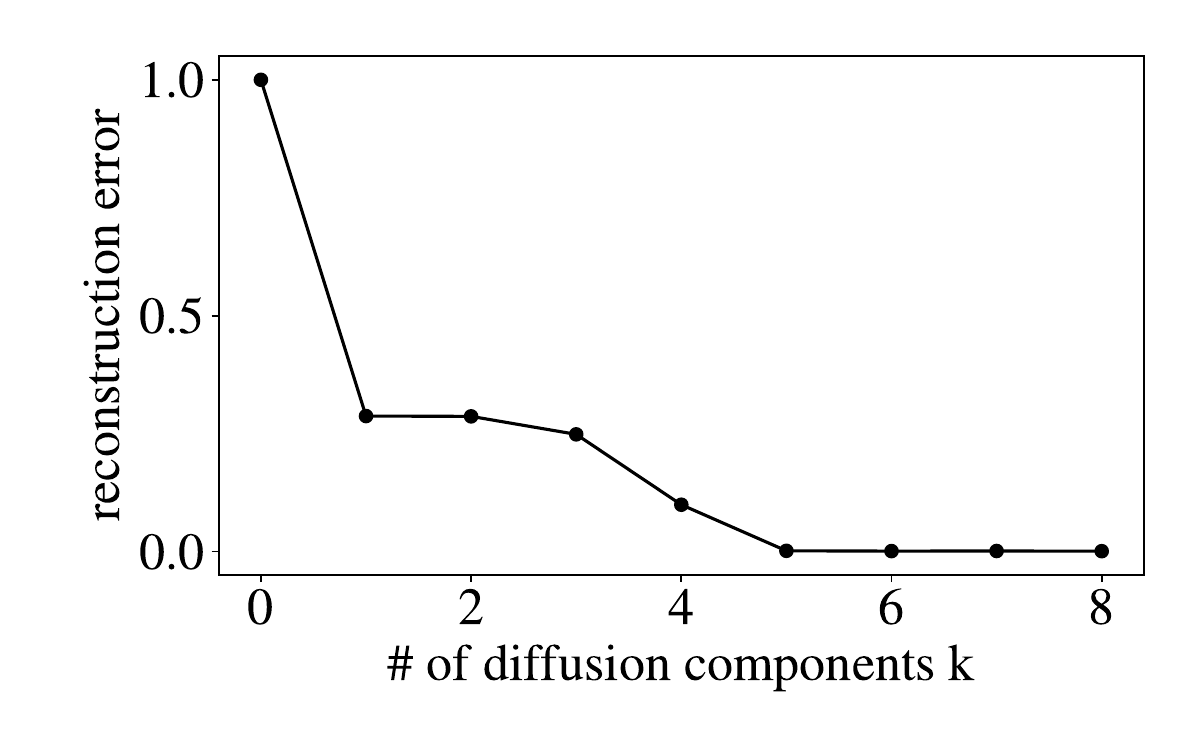}
    \caption{NRE of the Diffussion Map of the Swiss roll as a function of the number of diffusion components.}
    \label{fig:swissroll_nre}
\end{figure}

Figure \ref{fig:swissroll_nre} shows the NRE of the Diffusion Map of the Swiss roll, shown in Figure \ref{fig:swiss_roll}, as a function of the number $k$ of included diffusion components.
For the implementation details, we refer the reader to the appendix.

The NRE drops sharply when the first diffusion component is included, and reaches virtually zero only once the first five components are included.
This can be understood by looking at Figure \ref{fig:swiss_roll}, showing that the first component encodes the length direction and the fifth component encodes the width direction of the Swiss roll.
The first and fifth diffusion components together therefore provide a parametrization of the Swiss roll, allowing for near perfect reconstruction.
Diffusion components $\Psi_2$ through $\Psi_4$, on the other hand, are essentially functions of $\Psi_1$ (see Figure \ref{fig:swiss_roll}).
This is a peculiarity of the Diffusion Map, also noted by \cite{Nadler2007}, which occurs when the data manifold has directions of very unequal extent.
In our example, the Swiss roll has an aspect ratio of 90:21, which is why the smaller width direction is only represented in the fifth diffusion component.

\begin{figure}
    \centering
    \includegraphics[width=\linewidth]{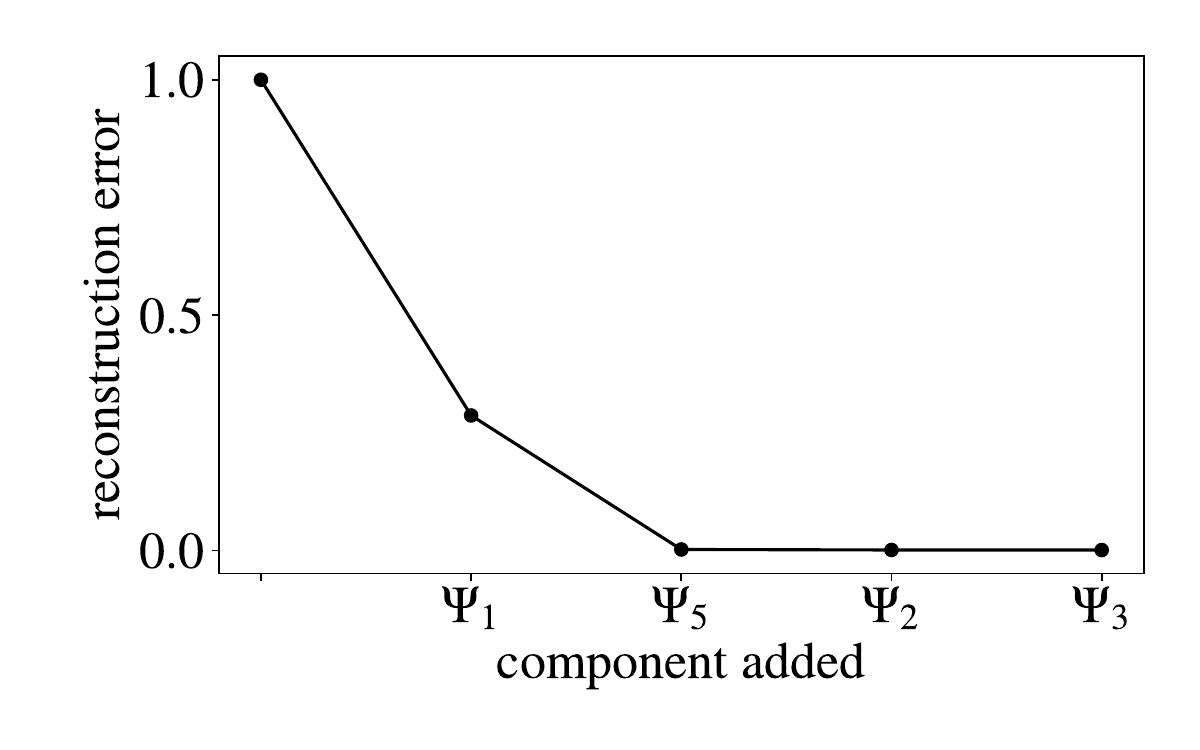}
    \caption{NRE of the Diffusion Map of the Swiss Roll, where components are added in the order $\Psi_1, \Psi_5, \Psi_2, \Psi_3, \ldots$}
    \label{fig:swissroll_narrow_nre_nc}
\end{figure}

The NRE applied to the example provided here shows that the best embedding is not necessarily given by the first $k$ consecutive diffusion components.
Other combinations can also be considered by the NRE, since it can be computed for arbitrary combinations of components.
Figure \ref{fig:swissroll_narrow_nre_nc} illustrates this. 
It again shows the NRE of the Diffusion Map of the Swiss roll, however, instead of in the usual order, components are added in the order $\Psi_1, \Psi_5, \Psi_2, \Psi_3, \ldots$, putting the two components which parametrize the manifold first.
% It shows that diffusion components $\Psi_1$ and $\Psi_5$ alone parametrize the data manifold, leading to almost lossless reconstruction.
Unlike in Figure \ref{fig:swissroll_nre}, the two-dimensional nature of the Swiss roll is immediately revealed, since the NRE reaches zero after including just two diffusion components.
This serves as an example that care must be taken when selecting the diffusion components.
% In principle, NRE can also enable a 

\section{Conclusion}

In this paper, we provided a first comprehensive review of the Diffusion Map method. We showed for the standard example of Swiss Roll data how parameter settings and preprocessing strategies (such as normalization, rescaling, and redundant variables) significantly influence the resulting manifold representation. Using a neural-network based method for quantifying the accuracy of the reduced data set, we highlight the importance of not simply taking the first few components but choosing the retained  Diffusion Map components carefully.

Our analysis shows that performance is strongly influenced by the Diffusion Map parameters setting and the presence (or not) of data pre-processing. We recommend to take time to test a wide range of parameters before taking the resulting embedding as reliable. Though the effect of the time parameter $t$ is negligible from a practical and qualitative perspective, special attention should be payed to the neighborhood parameters, $N$ and $\epsilon$. 
It must be tested carefully, keeping in mind the effect of each parameter and finding a balance between computational efficiency and choosing a neighborhood that is large or small enough to capture the relevant information without mixing unrelated structures. Testing the effects of density through $\alpha$ is also important, especially when density plays a significant role in the data of interest.
The effect of pre-processing is usually underestimated, but our observations confirm that it must be treated seriously. Users should be cautious in scenarios involving a mixture of discrete and continuous variables, scaling of variables or a certain redundancy among variables, which can lead to distortions in the resulting embedding.
Awareness of these factors is essential to avoid misapplications and misinterpretations.

Additionally, the %use of the 
Neural Reconstruction Error %analysis to find the optimal Diffusion Map embedding is a completely new and particularly useful tool. It
provides a quantitative, consistent and reliable criterion to select a suitable, accurate Diffusion Map embedding.%, something that we found previously missing in the literature and it will ease and refine the use of this method. 

Taken together, we provide clear guidance for applying the Diffusion Map in practice, highlighting the aspects that require the most careful attention and offering tools to make the most of this useful method. The study of large datasets is becoming increasingly common, and the proper use of methods that enable their dimensionality to be reduced is more relevant than ever.

\bibliography{literature.bib}

\end{document}